\documentclass[
twocolumn,
]{ceurart}

\usepackage{listings}
\usepackage{graphicx}
\newcommand{\ms}[1]{\tiny{$\pm$#1}}
\usepackage{spverbatim} 
\usepackage{hyperref}

\usepackage{booktabs} 
\usepackage{subfigure}
\usepackage{amssymb}
\usepackage{tabularx}
\usepackage{multirow}
\usepackage{makecell}
\usepackage[dvipsnames]{xcolor}
\usepackage{soul}

\begin{document}

\copyrightyear{2022}
\copyrightclause{Copyright for this paper by its authors.
  Use permitted under Creative Commons License Attribution 4.0
  International (CC BY 4.0).}

\conference{2nd Workshop on Perspectivist Approaches to NLP}

\title{Annotation Imputation to Individualize Predictions: Initial Studies on Distribution Dynamics and Model Predictions}

\author[1]{London Lowmanstone}[%
orcid=0000-0002-5457-6553,
email=lowma016@umn.edu,
]
\cormark[1]
\fnmark[1]
\address[1]{University of Minnesota}

\author[2]{Ruyuan Wan}[
email=rwan@nd.edu,
]
\fnmark[1]
\address[2]{University of Notre Dame}

\author[1]{Risako Owan}[
email=owan002@umn.edu,
]

\author[3]{Jaehyung Kim}[
email=jaehyungkim@kaist.ac.kr,
]
\address[3]{KAIST}

\author[1]{Dongyeop Kang}[
email=dongyeop@umn.edu,
]

\cortext[1]{Corresponding author.}
\fntext[1]{These authors contributed equally.}

\begin{abstract}
Annotating data via crowdsourcing is time-consuming and expensive. Due to these costs, dataset creators often have each annotator label only a small subset of the data. This leads to sparse datasets with examples that are marked by few annotators.
The downside of this process is that if an annotator doesn't get to label a particular example, their perspective on it is missed.
This is especially concerning for subjective NLP datasets where there is no 
single
correct label: people may have different valid opinions. Thus, we propose using imputation methods to generate the opinions of all annotators for all examples, creating a dataset that does not leave out any annotator's view. We then train and prompt models, using data from the imputed dataset, to make predictions about the distribution of responses and individual annotations. 

In our analysis of the results, we found that the choice of imputation method significantly impacts soft label changes and distribution.
While the imputation introduces noise in the prediction of the original dataset, it has shown potential in enhancing shots for prompts, particularly for low-response-rate annotators.
We have made all of our code and data publicly available.\footnotemark[1]
\end{abstract}

\begin{keywords}
    natural language processing \sep
    imputation \sep
    matrix factorization \sep
    content filtering \sep
    large language models \sep
    annotation \sep
    NLPerspectives \sep
    LeWiDi
\end{keywords}

\maketitle
\footnotetext[1]{https://github.com/minnesotanlp/annotation-imputation}
\section{Introduction}
Natural language processing (NLP) models rely on large amounts of data that is expensive and time-consuming to label \cite{sharir_cost_2020}. Crowdsourcing has emerged as a popular solution to this problem, but it comes with its own challenges, principal among them being annotator disagreement \cite{checco2017let, kairam2016parting}. Although there are many possible causes of disagreement, the common causes are annotator
subjective judgment and language ambiguity \cite{Uma2022ScalingAD}. Not taking into account the inherent subjectiveness and ambiguity of some instances can lead to inaccurate predictions \cite{fornaciari2021beyond}. Thus, in recent years, researchers have begun to recognize the importance of disagreement, advancing models and datasets that accurately reflect disagreement, rather than ignoring it or working around it \cite{leonardelli_semeval-2023_nodate}.

\begin{figure}[t]
\centering
\includegraphics[width=0.48\textwidth,trim={0.9cm 0.9cm 0.9cm 0.9cm},clip]{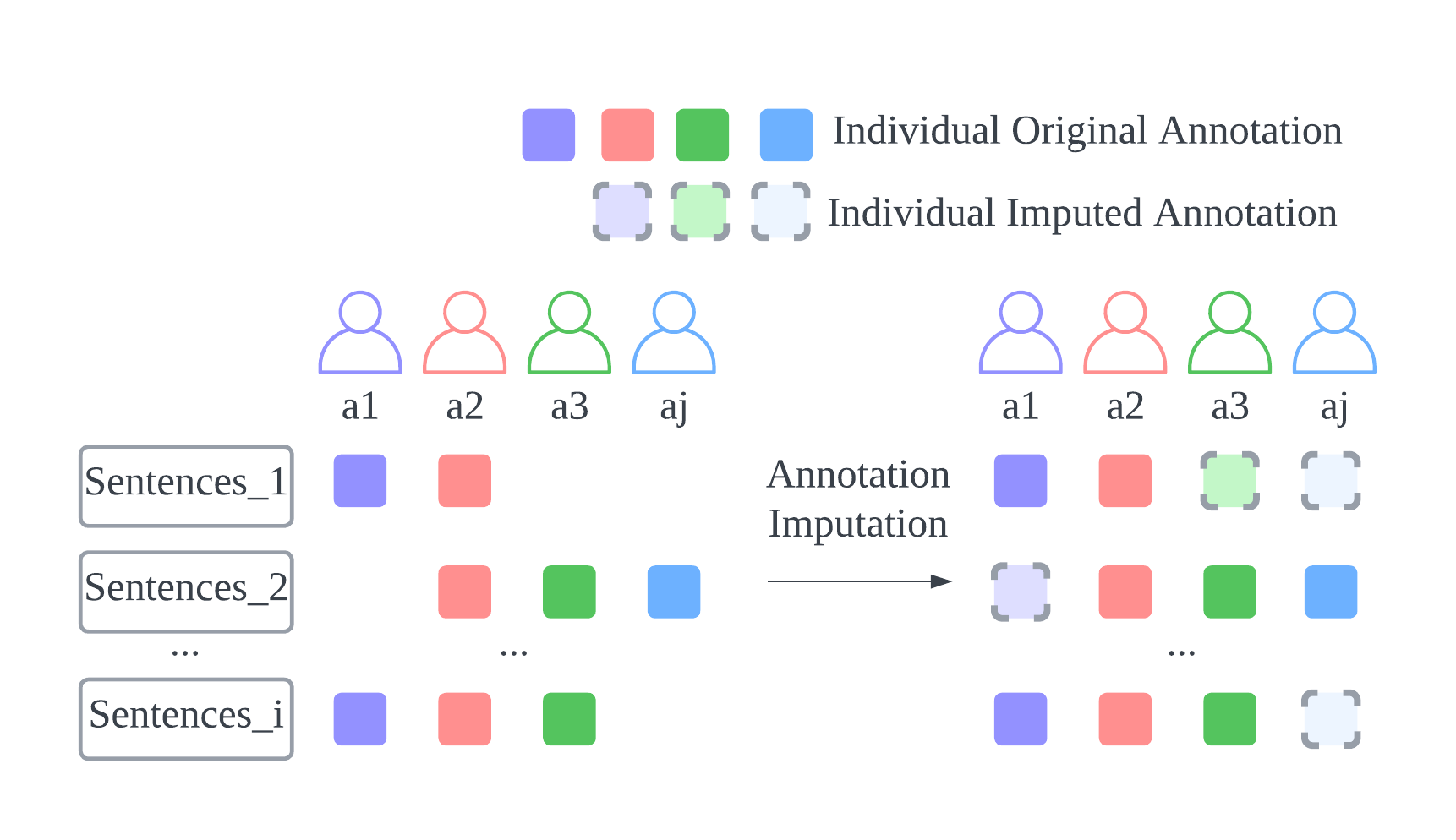}
\caption{Annotation imputation by using individualized prediction. Each square represents a single annotation. The original dataset on the left is missing some annotations from annotators. We then make predictions as to how each of the missing annotations would be filled in, resulting in the imputed dataset on the right. The slightly transparent squares indicate imputed annotations that are not in the original dataset. We then analyze how the imputed dataset on the right differs from the original data on the left.}
\label{fig:imputation_concept}
\end{figure}

In order for models to accurately reflect disagreement, they must accurately model true human populations. Here, we frame the problem of making accurate predictions for individual annotators as an \textit{imputation} problem: given a spreadsheet with rows corresponding to text and columns corresponding to annotators, how would one accurately fill in the spreadsheet in order to correctly predict how each annotator will label each piece of text? Figure \ref{fig:imputation_concept} visualizes this approach, which, ideally, enables dataset creators to generate additional annotations without extensive crowdsourcing.

We postulate that annotators who have historically assigned the same labels to identical text segments may, given similar contexts in unseen data, continue to demonstrate congruent labeling behavior.
Thus, imputation methods, which take in data containing all of the dataset annotations, should be able to discover patterns to relate annotators and annotations in order to make accurate predictions as to how a particular annotator might label a particular example, based on how other annotators labeled the same or similar examples.

Matrix factorization techniques used in recommendation systems and annotator-level models of disagreement both make predictions about individual annotations made by individual annotators. Thus, our analyses can be applied to both types of models in order to reveal differences between the original data and imputed data created by these models. In our work, we impute datasets by utilizing two matrix factorization methods, kernel matrix factorization and neural collaborative filtering, and a supervised learning model (Multitask) proposed by \cite{davani_dealing_2022}, that models disagreement at the annotations level \cite{rendle_online-updating_2008, he_neural_2017, davani_dealing_2022}. Through our analyses, we find that imputation greatly transforms the distribution of annotations (including lowering the variance of the data) and creates noticeable changes in examples' soft labels. 

After imputing and analyzing the data, we use the imputed datasets to train and prompt models that make individualized predictions. For training, we use the Multitask model from \cite{davani_dealing_2022} in order to make aggregate and individualized predictions and find that training on imputed data harms prediction performance. For prompting, we use GPT-3 (text-davinci-003) and ChatGPT (3.5-turbo) and provide the models with prompts containing either imputed or non-imputed data to determine their impact on the models' ability to make individualized and distributional label predictions. We find that adding prompt shots via imputation improves ChatGPT's performance for predicting annotations of low-response-rate annotators, but does not consistently improve other areas of prediction such as distributional label prediction, individualized prediction for high-response annotators, or merely replacing human annotations with imputed data \cite{brown_language_2020}.

In summary, our primary contributions are:
\begin{enumerate}
    \item Framing individualized prediction as an imputation problem
    \item Analysis techniques to compare imputed data to real data: 
    \begin{enumerate}
        \item \textit{Distribution Analysis}, which focuses on transformations of the underlying distribution of annotations after imputation. We show that different imputation methods significantly change the underlying annotation distributions.
        \item \textit{Soft Label Analysis}, which focuses on shifts in the soft label after imputation compared to the original data. 
        We provide a visualization technique for viewing how the soft labels change after imputation.
        \item \textit{Usage Analysis}, which focuses on how models perform after training on or being prompted with imputed data. We show that kernel matrix factorization, neural collaborative filtering, and Multitask imputation tend to harm the capabilities of Multitask and GPT models to make individual, soft-label, and aggregate predictions, except in the case of using imputation to increase the number of shots to prompts for making individualized predictions for low-response-rate annotators.
    \end{enumerate}
\end{enumerate}
\section{Related Work}

\paragraph{Disagreement in NLP}
Disagreement has been found within NLP datasets for many years \cite{poesio2005reliability, versley_vagueness_2008, wan-badillo-urquiola-2023-dragonfly, leonardelli_semeval-2023_nodate}.
However, recently, there has been much work done on developing and evaluating models that model disagreement within datasets, rather than ignore disagreement \cite{davani_dealing_2022, gordon_jury_2022,fornaciari2021beyond,gordon_disagreement_2021, wan2023everyone}.

In particular, the SemEval-2023 Learning with Disagreements (LeWiDi) task invites competitors to create models that predict soft labels of human disagreement for different text inputs \cite{leonardelli_semeval-2023_nodate}. While hard labels provide a definitive categorization for data points, soft labels offer a probabilistic interpretation, capturing the uncertainties or nuances in classification.
Multiple submissions for this task used models proposed by \cite{davani_dealing_2022} in order to make predictions at the individual level. Success at the task was determined by micro F1 score on gold labels and cross-entropy on soft-labels. Within the task, all dataset labels were binary, and no metric was used to measure success at the level of individual annotators.

The authors of \cite{uma_learning_2021} propose multiple different methods for evaluating models that make individualized predictions. Among these are Jensen-Shannon divergence, a symmetric variation of Kullback-Leibler (KL) divergence and cross-entropy. F1 score is also a proposed metric, but only for aggregate labels, not individual labels.

Another model for approaching disagreement is Jury Learning, where individuals' annotations are modeled in order to form ``juries" of different demographics \cite{gordon_jury_2022}. In their paper, the authors analyze how using data generated by ``juries" affects the aggregate label, particularly in the case of contentious texts \cite{gordon_jury_2022}.

\begin{figure*}[t]
    \centering
    \includegraphics[width = 1\linewidth]{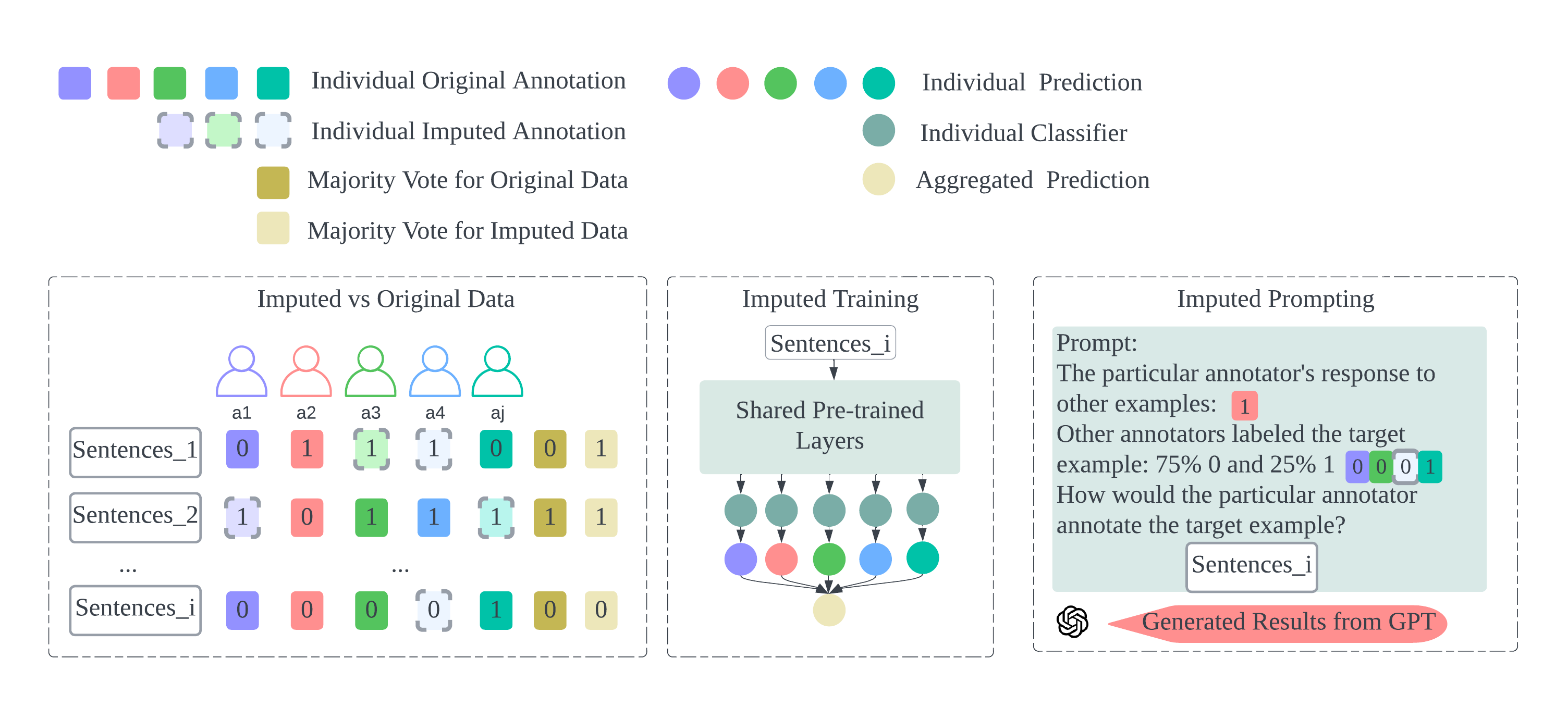}
    \caption{Three experiments of annotation imputation: (1) comparing imputed vs original data, (2) training on imputed data (3) generating with prompts based on imputed data.}
    \label{fig:pipeline}
\end{figure*}

\paragraph{Collaborative Filtering in Recommendation Systems}
Similar to modeling disagreement, collaborative filtering systems also create individualized predictions of human behavior in order to make relevant recommendations. Contrary to disagreement models in natural language processing, these systems are entirely dependent on user-provided annotations and lack the ability to predict the reactions of new users to unseen text.

When evaluating performance of collaborative filtering systems, metrics are generally focused on accuracy of predictions, rather than quantifying and visualizing changes in the distribution of data \cite{herlocker_evaluating_2004, isinkaye_recommendation_2015}. These metrics provide good signals for the success of a model, but do not help with understanding how models modify data when they do not match the original data.
\section{Annotation Imputation For Individualized Predictions}
First, we compared how various imputation methods handle and fill in the missing annotations. We then trained supervised models and used GPT-based prompting to evaluate imputation's impact on aggregate and individualized prediction.

\subsection{Annotation Imputation}
In order to understand how individualized prediction affects data, we use three different methods: kernel matrix factorization, neural collaborative filtering (NCF), and a Multitask supervised neural network model from \cite{davani_dealing_2022}.\footnote{The hyperparameters used for each of the models can be found in Appendix \ref{app:hyperparameters}.}

Kernel matrix factorization relies on kernels to project data to a higher dimensional space where more complex patterns can be found in order to generate a matrix factorization which is used for imputation. NCF matrix factorization relies on neural networks rather than kernels to compute a matrix factorization of the data, and Multitask relies purely on neural networks to make individualized predictions. All methods employ a core process: identifying patterns between annotators and annotations across the dataset.

For our experiments, kernel matrix factorization is implemented primarily using off-the-shelf code \cite{do_matrix_2022}. In addition, we add a grid search component which determines the best model hyperparameters by holding out 5\% of the given training data as validation data, and choosing the hyperparameters that resulted in the lowest RMSE score on the validation data. See Appendix \ref{app:hyperparameters} for details.

Neural collaborative filtering was implemented based on the work of \cite{he_neural_2017}. The details of our implementation can be found via our code. For this model, we also use an additional grid search component which determines the best model hyperparameters. However, we choose the hyperparameters for this model based on the lowest RMSE score when evaluated on all training examples, rather than a held-out validation set. See Appendix \ref{app:hyperparameters} for details.

\subsection{Imputed Training}
In this stage, we use the Multitask model from \cite{davani_dealing_2022} on both original and imputed data and compare the evaluation results in order to understand how imputed data impacts model training. We follow a similar setup to \cite{davani_dealing_2022} by using 5-fold validation and averaging the results across the folds \cite{davani_dealing_2022}. However, in order to account for dataset imbalance in our datasets, we report weighted F1 scores, rather than macro F1 scores. Note that the data from each validation fold is hidden from the imputer, so as not to cause data leakage. Details of the model's architecture can be found in Appendix \ref{app:multitask-model}, and hyperparameter details can be found in Appendix \ref{app:hyperparameters}. The same model is used both for imputation and training (see Section \ref{sec:experiments}).

\subsection{Imputed Prompting}
We also conducted three key experiments using GPT-3 (text-davinci-003) and ChatGPT (3.5-turbo) to better understand the impact of imputation on predictions made by GPT-based models \cite{brown_language_2020}:
The first experiment tests the impact of using imputed data when making individualized predictions for low-response-rate annotators. 
The second experiment makes individualized predictions for all annotators (not just low-response-rate annotators), but also adds original distribution information, imputed distribution information, or the original majority-voted label near the end of the prompt in addition to the included individual examples to quantify the impact of the extra information on predictions. 
The third tests individualized predictions when either original or imputed data from three distinct annotators is provided in the prompt. 
Of these, imputation only had a positive impact on making individualized predictions for low-response-rate annotators; the other two experiments are included in Appendix \ref{app:chatgpt-extra-results}.

For all experiments, we create prompt skeletons, which are then filled in with data and/or text, depending on the experiment run (see Appendix \ref{app:prompts}). This enables us to understand the influence of different prompts and data.

\paragraph{Individualized Predictions for Low-Response-Rate Annotators} In this experiment, we first isolated from each dataset the 30 annotators with the lowest number of annotations in the dataset. We then generated a prompt for each of those 30 annotators. Each prompt consists of at most 30 sentences and annotations from that annotator (if there were more, we discarded the extras and chose one to hold out, and if there were less, we included all but one to hold out). Following the real examples, we also included an additional 30 examples whose sentences are from the dataset (and differ from the previous 30 examples and the held-out example), but whose annotations are imputed via NCF. The final section of the prompt then asks ChatGPT to predict the annotator's annotation on the held-out example.

In the experiment, we test for differences between three different conditions:
\begin{enumerate}
    \item Including both the original and imputed data 
    \item Only including the original data 
    \item Only including the imputed data 
\end{enumerate}

In each of these conditions, outputs are considered correct if, after removing whitespace, they only contain the correct label. We conducted initial studies to discard particularly low-performing skeletons and infills. The remaining skeletons and infills are used for all conditions. (Details are provided in our code.) We then measure success of a condition based on the highest weighted F1 score achieved by a prompt skeleton within that condition.

\section{Experiments}\label{sec:experiments}

\begin{table*}[t]
    \centering
    \small
    \begin{tabular}{@{}c@{}c@{}c@{}c@{}c@{}}
    \toprule
    \textbf{Dataset} & {\textbf{\#} \textbf{instances}} &{ \textbf{\#} \textbf{annotators}} &{ \textbf{\#} \textbf{annotation}} & \textbf{Label Types} \\
    \midrule
        \makecell{SChem \\ SChem5Labels} & \makecell{400 \\8007}& \makecell{100\\102}& \makecell{50\\5}& 
        \makecell{ No one believes (0), occasionally believed (1),\\
        controversial (2), common belief (3), universally true (4) }\\
    \midrule
        SBIC & 45223 & 304 & 3 &
        \makecell{Not offensive (0), maybe (0.5), offensive (1) }\\
    \midrule
        GHC & 27538 & 18  &  3-4 &\makecell{Not hate speech (0), hate speech (1)}\\
    \midrule
        Sentiment & 14070 & 1481 & 4-5& \makecell{Very negative (-2), somewhat negative (-1), neutral (0), \\ somewhat positive (1), very positive (2)} \\
    \midrule
        Politeness & 4338 & 219 & 5 & \makecell{A scale from polite (1) to impolite (25). } \\
        
    \bottomrule
    \end{tabular}
    \caption{Statistics and label information on the six datasets we use across our analyses. The statistics include the number of unique text instances, the number of unique annotators, and the number of annotations per text instance in the six datasets.}
    \label{tab:dataset}
\end{table*}

Our experiments involve: (1) comparing imputed and original data, (2) conducting training using imputed data, and (3) prompting generation based on imputed data, all illustrated in Figure \ref{fig:pipeline}.
\paragraph{Datasets}
In order to ensure a diversity of data, we utilize six different datasets in our analysis: Social Chemistry (SChem) \cite{Forbes2020SocialC1}, Social Bias Inference Corpus  (SBIC) \cite{Sap2020SocialBF}, Gab Hatespeech Corpus (GHC) \cite{kennedy2018introducing,kennedy2022introducing}, Sentiment dataset \cite{Diaz2018AddressingAB}, and Politeness dataset \cite{DanescuNiculescuMizil2013ACA}. Additionally, we isolate examples from the SChem dataset that were labelled by 5 annotators in order to form the SChem5Labels dataset. Our datasets are summarized in Table \ref{tab:dataset}, and more details can be found in Appendix \ref{app:datasets}.

\subsection{Imputed vs Original Data}
\label{sec:imputed-vs-original}

\paragraph{Imputation}
We impute each of the datasets with each of the imputation methods. However, in order to judge which methods have the best performance, we also test imputing the data while withholding 5\% of the annotations for evaluation. Withheld data is chosen in a manner that reduces duplicate examples and annotators within the withheld data in order to provide a more diverse test set (details can be found in our code).

Table \ref{tab:rmse} summarizes the RMSE score for each of the methods on each of the datasets when evaluated on the withheld data. Note that the
Politeness dataset collects labels ranging from 1 to 25, implying a broader variance compared to other datasets. Consequently, RMSE values are expected to be higher for the Politeness dataset. 
We also find that while Multitask and NCF perform best on different datasets, kernel matrix factorization is never the best method, and is in fact always dominated by the NCF method.

\begin{table*}[t]
	\begin{center}
	\begin{tabular}{@{}r|cccccc@{}}
		\toprule
		Method & SChem & SChem5Ls & GHC & SBIC & Sentiment & Politeness
		      \\ 
                \midrule
        Multitask
                & 0.82
                & \underline{0.72}
                & \textbf{0.32}
                & \textbf{0.64}
                & 1.14
                & 4.41
               \\
	  NCF     & \textbf{0.63}
                & \textbf{0.66}
                & \underline{0.35}
                & \underline{0.65}
                & \textbf{0.90}
                & \textbf{3.69}
               \\
		Kernel
                & \underline{0.71}
                & \underline{0.72}
                & 0.36
                & 0.90
                & \underline{1.03}
                & \underline{4.39}
               \\
        \bottomrule
	\end{tabular}
    \end{center}
    \caption{RMSE scores of the different imputation methods across datasets. All models were run once except kernel matrix factorization, whose reported scores are the median of 3 runs with differing random seeds. The lowest RMSE score on each dataset is in \textbf{bold}, and the second-lowest is \underline{underlined}.}
     \label{tab:rmse}
\end{table*}

After the data is imputed, we use two analyses in order to better understand how imputed data differs from original data.

\paragraph{Distributional Analysis}
\begin{figure*}[h]
    \centering
    \includegraphics[width = 1\linewidth]{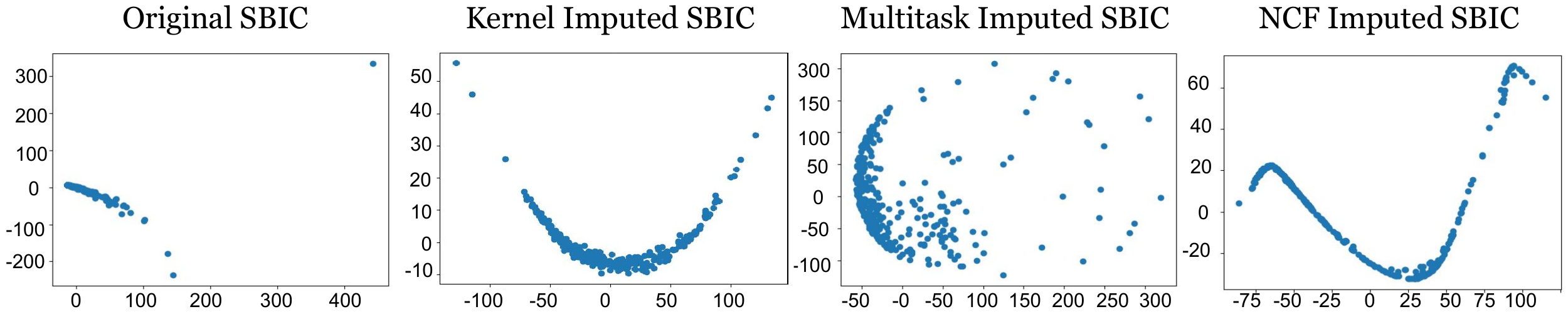}
    \caption{PCA projection of the SBIC dataset before and after using imputation.}
    \label{fig:pca}
\end{figure*}

The first analysis (distribution analysis) applies principal component analysis (PCA) to both imputed and original data to visualize shifts in the distribution of example ratings. In order to apply PCA, we represent each text as a vector of its annotations, where missing annotations are filled in with a value of 10, which is far outside the range of valid annotation labels for these datasets \cite{roy_all_2019}. We also calculate the change in variance between imputed and original data, and graph this variance against the disagreement rate across examples. The disagreement rate is computed as the number of annotations that disagree with the majority-voted label for that example, divided by the total number of annotations for that example. The majority-voted label for imputed data is computed on the imputed data.

When we project the annotations to two dimensions using PCA, we find that different imputation methods cause significant changes to the distribution of the data as shown in Figure \ref{fig:pca}.\footnote{Other datasets' results can be found in Appendix \ref{app:pca}.} Each imputation method generates an extremely different underlying distribution for the annotations.

\begin{figure}[t]
    \centering
   \includegraphics[width=1.1\linewidth]{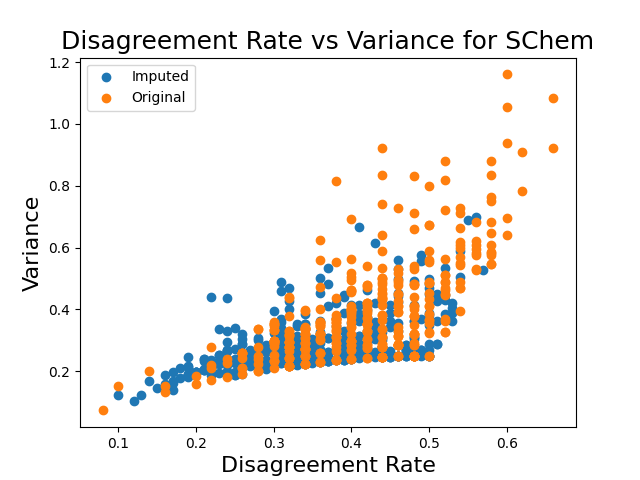}
    \caption{A graph displaying how the variance has decreased after using NCF matrix factorization. Each point represents an example. Variation is across annotations for that example, and disagreement rate is the percentage of people who disagree with the majority-voted annotation.}
    \label{fig:variance-and-disagreement}
\end{figure}

\begin{table*}[t]
	\begin{center}
	\begin{tabular}{r|cccccc}
		\toprule
		Statistic & SChem & SChem5Labels & Politeness & Sentiment & SBIC & GHC   
		      \\ 
                \midrule
	Avg Change: Variance 
                & \textbf{-0.096}
                & \textbf{-0.088}
                & \textbf{-6.987}
                & \textbf{-0.312}
                & \textbf{-0.044}
                & \textbf{-0.004}
               \\
        Avg Change: Disagreement Rate 
                & \textbf{-0.061}
                & \textbf{-0.060}
                & 0.18
                & \textbf{-0.106}
                & 0.044
                & \textbf{-0.006}
               \\
        \bottomrule
	\end{tabular}
    \end{center}
    \caption{Change in average variance and disagreement rate due to using NCF matrix factorization to impute the dataset. Instances where the variance or disagreement rate are \textbf{lowered} due to imputation appear in bold.}
     \label{tab:variance-and-disagreement}
\end{table*}

In addition, we compute how the variance and disagreement rate change after imputation with NCF matrix factorization. Our results are compiled in Table \ref{tab:variance-and-disagreement}, and we also provide Figure \ref{fig:variance-and-disagreement} to display the results on the SChem dataset.
Results from other methods can be found in Appendix \ref{app:variance-and-disagreement}. 
Across all datasets, we find that imputation decreases variance, indicating that NCF matrix factorization does not accurately model the diversity of human annotations. We can observe this lowered variance in both Figure \ref{fig:pca} and Figure \ref{fig:variance-and-disagreement} by comparing the scale of the plots in the PCA visualization and by comparing the heights of the points in the variance plot. We also find that NCF matrix factorization tends to, but does not always, lead to more agreement with the majority-voted annotations.

Overall, the chosen method for individualized prediction has a large impact on the structure underlying the predictions, even within the same dataset. We also find imputers can lower variance and raise agreement within the dataset, demonstrating that imputation models may not always capture the diversity and disagreement of real human annotators.

\paragraph{Soft Label Analysis}
\label{sec:results:soft-label-analysis}
The second analysis visualizes differences between soft labels of examples between the original and imputed data. To create the visualization, we assign each label to a color and then generate horizontal bars of equal size where the proportion of the bar containing that color corresponds to the proportion of annotations with that label. This enables us to directly compare how different imputation methods alter the soft label distribution. Similar to \cite{uma_learning_2021}, we also calculate the Kullback-Liebler (KL) divergence between the original distribution of data and the imputed data in order to numerically quantify the difference between distributions.

\begin{figure*}[h]
    \centering
    \includegraphics[width=0.99\linewidth]{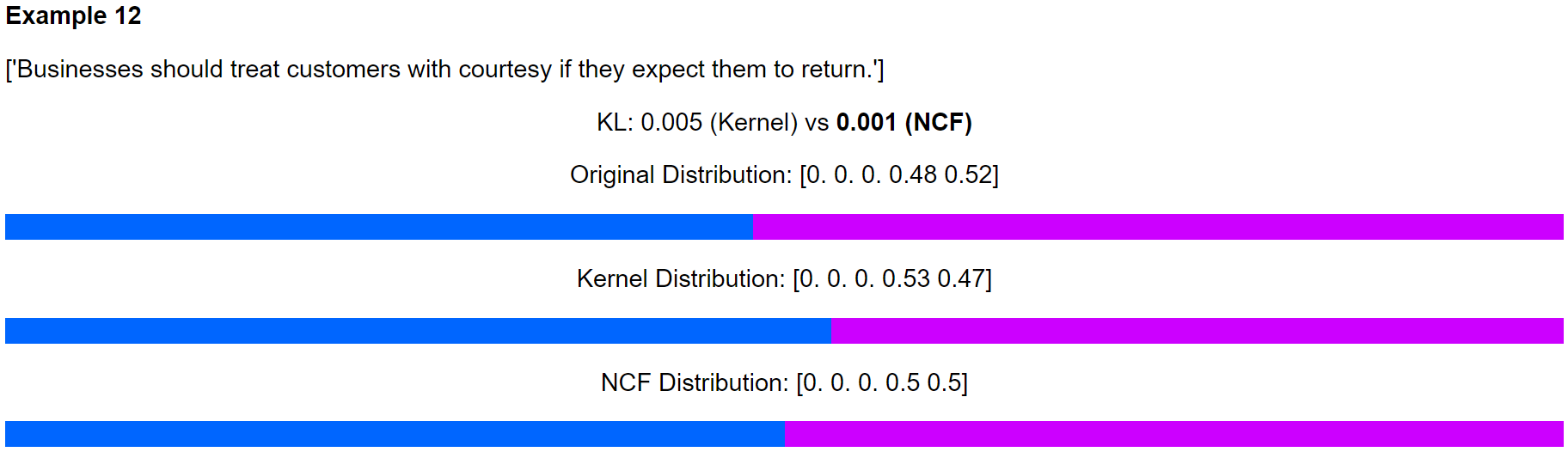}
    \includegraphics[width=0.99\linewidth]{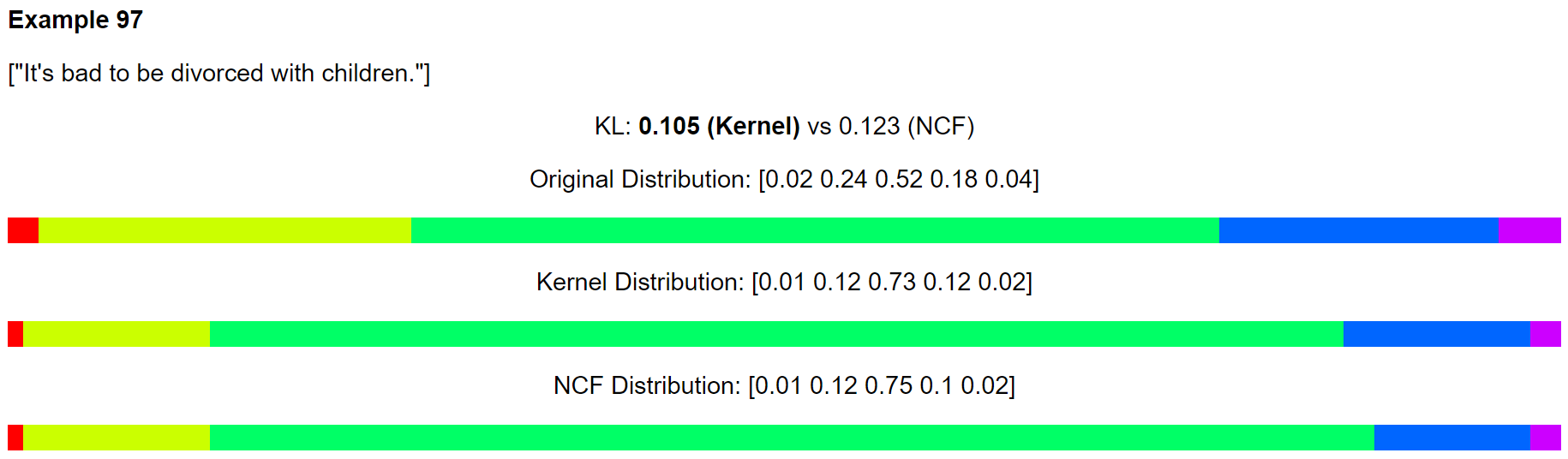}
    \caption{Visualizations showing the shift in distribution between the original distribution (soft label), the distribution after using kernel matrix factorization, and the distribution after using NCF matrix factorization on examples within the SChem dataset. The exact proportions of each label are listed, as well as the KL divergence score for each method at the top, with the method best reflecting the original distribution in \textbf{bold}. The top Example 12 illustrates an example for which both NCF and kernel matrix factorization do a good job of keeping the original soft-label, whereas for the bottom Example 97, both methods highly over-estimate the proportion of annotators who respond with the middle label.}
    \label{fig:imputation-website}
\end{figure*}

Through our soft label analysis, we find that different imputation methods lead to varying changes of the soft label of examples after imputation. Figure \ref{fig:imputation-website} demonstrates how imputation changes the distribution of the data for a given example and allows one to directly compare different imputation methods to see how they modify the data. In the case of Figure \ref{fig:imputation-website}, we see an example from the SChem dataset which shows that kernel matrix factorization predicted a much smaller proportion of annotators to give the highest rating (in pink) than was in the original dataset, while NCF matrix factorization predicted a moderately larger proportion of users to give the second-highest rating (in blue) than the original dataset.

Since we are interested in understanding how these soft labels differ from the original data, we also compute the KL divergence between the imputed data and the original data. Note that if one would like a symmetric metric, the Jensen-Shannon divergence could be computed here as well \cite{uma_learning_2021}. In the particular example in Figure \ref{fig:imputation-website}, we see that the KL divergence score on Example 97 for Kernel is 0.105, compared to the 0.123 divergence score by NCF. We provide a selection of multiple examples in Appendix \ref{app:disagreement-website}. We also provide the average and standard deviation of KL divergence from the original data for each dataset and each imputation method in Table \ref{tab:kl-divergence}. Overall, we see that NCF matrix factorization tends to best preserve the soft label of the original dataset when compared to kernel matrix factorization and the multitask model, as it is always either best or second-best. However, performance is dataset-dependent, and kernel and multitask achieve the best fidelity to the original soft labels for the Sentiment and SBIC datasets, respectively.

Overall, soft labels do not remain consistent through imputation, and some methods of individualized prediction may tend to better preserve soft labels than others. In our case, NCF matrix factorization best preserved the soft labels.

\begin{table*}[t]
	\begin{center}
    \scalebox{1}{
	\begin{tabular}{@{}r|cccccc@{}}
		\toprule
		Method / Dataset & SChem & SChem5Ls & Sentiment & SBIC   
		      \\ 
                \midrule
	  NCF     & \textbf{0.030\ms{0.0262}}
                & \textbf{0.233\ms{0.305}}
                & \underline{0.582\ms{0.678}}
                & \underline{0.413\ms{0.615}}
               \\
		Kernel
                & \underline{0.036\ms{0.0268}}
                & \underline{0.292\ms{0.363}}
                & \textbf{0.709\ms{0.700}}
                & 0.465\ms{0.654}
               \\
        Multitask
                & 0.080\ms{0.109} 
                & 0.317\ms{0.369}
                & 0.540\ms{0.359}
                & \textbf{0.307\ms{0.359}}
               \\
        \bottomrule
	\end{tabular}
    }
    \end{center}
    \caption{Average and standard deviation of the KL divergence across datasets and individualized prediction methods. The method which best preserved the original distribution is in \textbf{bold}, and the KL divergence of the second-best method is \underline{underlined}. Note that preserving the soft label / distribution is not necessarily indicative of accuracy or performance.}
     \label{tab:kl-divergence}
\end{table*}

\begin{table}[t]
	\begin{center}
	\begin{tabular}{@{}r|cccccc@{}}
		\toprule
		Method / Dataset & Politeness & GHC & SChem   
		
		              \\ \midrule
	  Original     & \textbf{0.33\ms{0.004}}  
                        & \textbf{.89\ms{0.003}}  
                        & \textbf{0.53\ms{0.009}} 
		               \\
                            \midrule
        NCF
                        & {0.18\ms{0.005}}
                        & \underline{0.88\ms{0.013}}  
                        & {0.52\ms{0.014}} 
                          \\
        Multitask
                        & \underline{0.31\ms{0.007}}  
                        & \textbf{0.89\ms{0.009}}
                        & \underline{0.52\ms{0.006}}  
		               \\
                 \bottomrule
	\end{tabular}
    \end{center}
    \caption{Average weighted F1 score of \textit{individualized} predictions made by the Multitask classifier trained on data generated by either NCF matrix factorization or a separate Multitask model. All the values and error bars are mean and standard deviation across five folds. The best and the second best results on each dataset are indicated in \textbf{bold} and \underline{underline}, respectively.}
     \label{tab:training_results_individual}
\end{table}

\begin{table}[t]
	\begin{center}
    \scalebox{1}{
	\begin{tabular}{@{}r|cccccc@{}}
		\toprule
		Method / Dataset & Politeness & GHC & SChem   
		
		              \\ \midrule
	  Original     & \textbf{0.38\ms{0.019}}  
                        & \textbf{.919\ms{0.004}}  
                        & \textbf{0.611\ms{0.91}}
		               \\
                            \midrule
        NCF
                        & {0.22\ms{0.007}}
                        & {0.912\ms{0.010}}  
                        & \textbf{0.611\ms{0.91}}
                          \\
        Multitask
                        & \underline{0.34\ms{0.016}}  
                        & \underline{0.915\ms{0.009}}
                        & \textbf{0.611\ms{0.91}}
		               \\
                 \bottomrule
	\end{tabular}
    }
    \end{center}
    \caption{Average weighted F1 score of \textit{aggregate} (majority-voted) predictions made by the Multitask classifier trained on data generated by either NCF matrix factorization or a separate Multitask model. All the values and error bars are mean and standard deviation across five folds. The best and the second best results on each dataset are indicated in \textbf{bold} and \underline{underline}, respectively.}
     \label{tab:training_results_aggregate}
\end{table}

\begin{table*}[h]
  \centering
  \begin{tabular}{|c|ccc|ccc|}
    \hline
    \multirow{2}{*}{\textbf{Dataset}} & \multicolumn{3}{c}{\textbf{Not Imputed}} \vline & \multicolumn{3}{c}{\textbf{Imputed}} \vline \\
    & \textbf{Disagreement} & \textbf{N} & \textbf{Value} & \textbf{Disagreement} & \textbf{N} & \textbf{Value} \\
    \hline
    Politeness & Low & 1306 & \textbf{0.481\ms{0.015}} & Low & 1306 & \underline{0.352\ms{0.024}} \\
    & Medium & 2267 & 0.293\ms{0.013} & Medium & 2267 & 0.203\ms{0.008} \\
    & High & 762 & 0.193\ms{0.013} & High & 762 & 0.121\ms{0.005} \\
    \hline
    GHC & Low & 20344 & \underline{0.965\ms{0.004}} & Low & 20344 & \textbf{0.968\ms{0.003}} \\
    & Medium & 814 & 0.721\ms{0.012} & Medium & 814 & 0.654\ms{0.027} \\
    & High & 6392 & 0.717\ms{0.006} & High & 6392 & 0.654\ms{0.021} \\
    \hline
    SChem & Low & 133 & \textbf{0.608\ms{0.043}} & Low & 133 & \underline{0.604\ms{0.050}} \\
    & Medium & 137 & 0.590\ms{0.033} & Medium & 137 & 0.581\ms{0.030} \\
    & High & 130 & 0.404\ms{0.050} & High & 130 & 0.394\ms{0.045} \\
    \hline
  \end{tabular}
  \caption{F1 values from individualized prediction done by the Multitask model, broken out by disagreement in the original dataset. The highest F1 score for each dataset is in \textbf{bold}, and the second highest is \underline{underlined}. The ``N" column signifies how many examples are in each category.}
  \label{tab:training_results_disagreement}
\end{table*}

\begin{table*}[h]
	\begin{center}
    \scalebox{1}{
	\begin{tabular}{@{}r|cccccc@{}}
		\toprule
		Method / Dataset & Politeness & GHC & SChem & SChem5L & SBIC & Sentiment
		              \\ \midrule
        Combined
                        & 0.13 & 0.75 & 0.50 & \underline{0.60} & \textbf{0.95} & \underline{0.58}
		               \\
	  Original        & \textbf{0.14} & \underline{0.85} & \underline{0.53} & 0.49 & \underline{0.93} & 0.31
		               \\
        Imputed
                        & 0.07 & \textbf{0.86} & \textbf{0.56} & \textbf{0.65} & \textbf{0.95} & \textbf{0.60}
                          \\
                 \bottomrule
	\end{tabular}
    }
    \end{center}
    \caption{Highest Weighted F1 score for predicting the annotations of 30 users with the lowest response rate in the dataset across multiple prompt skeletons and infills of those skeletons. Imputation is done via the NCF method. The best result for a dataset is in \textbf{bold}, while the second-best is \underline{underlined}.}
    \label{tab:low-response-results}
\end{table*}

\subsection{Imputed Training}
\label{sec:imputed-training}

For imputed training, we train the Multitask model on the original data, data imputed by NCF, and data imputed by a separate Multitask model. (Since RMSE scores from kernel matrix factorization are worse than NCF on each dataset, we omit kernel matrix factorization from this experiment.) After training the Multitask model on the original and imputed data, we report the average and standard deviation of the weighted F1 score for individual and aggregate predictions over 5 folds using 5-fold validation in Table \ref{tab:training_results_individual} and Table \ref{tab:training_results_aggregate}.
We observe that training on imputed data from NCF and Multitask results in performance worse than if we had used just the original data. This indicates that the predictions made by each of the methods biases the data in a way that does not match the true predictions that the annotators would have made.

However, not all prediction models had the same level of performance, and different datasets observed different results. Generally, using the original data resulted in the best outcomes, followed by using the Multitask model to impute the training data. Using NCF matrix factorization to impute the data resulted in the lowest performance.

When we break out the model's performance to examine success on examples with differing levels of disagreement (Table \ref{tab:training_results_disagreement}), we see that the model tends to perform much better on examples with higher agreement among annotators. We also see that the drop in performance from imputing data is fairly consistent across disagreement levels, except for the GHC dataset anomaly on low disagreement examples, where imputation helped slightly. How disagreement levels are computed is discussed in detail in Appendix \ref{app:disagreement-levels}.

Overall, this indicates that different methods of individualized predictions can introduce different biases into the data that cause methods trained on these predictions to perform worse than if they had trained on just the original data. Since we expect performance to increase with the amount of data provided, we conclude that these particular methods of individualized prediction likely introduce strong biases that do not reflect reality \cite{kaplan_scaling_2020}.

\subsection{Imputed Prompting}
\label{sec:imputed-prompting}

Here, we highlight the results of using imputed data to improve individualized predictions on low-response-rate annotators, as shown in Table \ref{tab:low-response-results}. 
(As mentioned above, other experiments are detailed in Appendix \ref{app:chatgpt-extra-results}) 
From the data, we observe that using solely imputed data outperforms using original data or adding original data to the imputed data for all datasets except for Politeness.

Politeness is likely an outlier due to the high range of potential labels in the Politeness dataset, leading the NCF method to impute labels that are unlikely to occur in the real dataset, causing ChatGPT to also predict unlikely labels. However, when the amount of labels is smaller (5 or less), using only imputed data increases performance.

We conjecture that since low-response-annotators in these datasets generally have far less than 30 original annotations, imputation enables us to provide more shots to ChatGPT than the original dataset could provide, thus enabling more accurate predictions than can be made without imputation. While more data is needed to determine why combining both imputed and original data performs poorly, we provide supporting experiments in Appendix \ref{app:ablation} to demonstrate that the performance improvement from using imputed data 
is particular to low-response-rate annotators and is caused by the imputed data, not the prompt text.
\section{Discussion}
Our analyses shed light on the impact of various imputation methods on the structure, soft label, and training/prompting viability of imputed data in the context of NLP annotation tasks in comparison to purely human-labeled data. We demonstrate that different imputation methods can lead to significantly different underlying distributions of the data, which can, in turn, affect the performance of models trained on this data. 
Furthermore, while imputation can introduce noise, diminishing the accuracy of predictions for the original dataset, it is essential to consider that the original dataset may not wholly capture the full spectrum of reality due to the absence of some annotator opinions.
This has important implications for the design and evaluation of individualized prediction models in various applications, as well as for understanding and quantifying the biases that may be introduced by such models.

Each one of our analyses focuses on a particular area of interest, which, together, help researchers and practitioners to better understand the predictions made by individualized prediction models. The distribution analysis provides information to those who are interested in ensuring that their model's predictions match the distribution of the original data and tools for analyzing changes in disagreement and variation. For those who are interested in soft labels, such as competitors in future LeWiDi tasks, our visualization helps with understanding how models estimate the soft label and computational tools for determining which models mimic the original soft label best. As we see a rise in human-level predictions from systems, it is important to understand if models can be trained or prompted with data created by individualized prediction models. We provide analyses from base systems indicating how the chosen imputation method may affect performance. Regardless of the scenario, our provided analyses enable researchers and practitioners who use models that make individualized predictions to better understand the differences between their model's predictions and real human annotations.

\section{Future Work}
While we include two different matrix factorization methods from collaborative filtering, content-based recommendation systems also provide individualized predictions, so future work includes applying our methods to a content-based recommendation system.

Also note that each of the methods we use is not state-of-the-art in their respective field. We have chosen baseline models for ease of implementation. Future work includes running our methods on more advanced systems that may make more accurate predictions.

We also have not conducted a user study to verify and quantify that our analysis methods help with understanding how predicted data differs from original data. Our analysis here is based on the fact that previous methods rely on aggregate metrics and do not provide fine-grained and comparative data between original and predicted data. Conducting a user study would allow us to provide explicit evidence of the exact amount of improvement our methods provide in general for understanding how individualized prediction impacts data.
\section{Limitations}
While our methods are extendable to any model that makes individualized predictions, we only test our methods on baseline models for both disagreement modeling and collaborative filtering. Thus, when
used on state-of-the-art methods, our methods may give very different results. However, we still expect these methods to be useful for understanding how imputation modifies the underlying data, even if those modifications do not match our results.

\section{Conclusions}
We have proposed and utilized four different methods of understanding how the predictions made by individualized prediction models differ from the original data. We found that for kernel matrix factorization, and NCF matrix factorization, the original soft label for the data shifts in different ways based on the method used, the variance in labels is overall lowered, and training on data created by these methods results in generally worse prediction performance, while imputed data can be used to increase the number of shots in prompts.

Overall, we hope that our analysis methods for models that make individualized predictions are applied to future models in order to help researchers and practitioners to better understand how their models' predictions differ from real human annotators.
\section*{Ethics Statement}
Any methods which attempt to make individualized predictions carry the risk of learning how to replicate aspects of individuals' identities in order to make better predictions. This may be viewed as data misuse, a violation of privacy, or a violation of the right to be forgotten.

Furthermore, there's an inherent ethical challenge in the goal of generating synthetic perspectives and opinions. The ability to synthetically generate opinions might inadvertently discourage practitioners from seeking real human input. This poses two primary risks: 1) it may lead to erroneous assumptions based on the synthetic data rather than actual human sentiments, and 2) it might marginalize authentic human participation, thereby weakening the quality and inclusivity of dataset and model development.

While our methods are designed to help detect when models may be incorrectly predicting human behaviors,
they are most effective
when applied to models performing imputation. Thus, advocating for the success of this work may inadvertently promote the creation and usage of models with the ethical concerns described above.

We urge creators of individualized prediction systems to always obtain consent from their users before applying models to their data and to maintain open and consistent communication about how their data may be used. We also advocate for a balanced approach, ensuring that while we progress in model development, real human perspectives remain at the core of our datasets and models.

\appendix
\clearpage
\section{Multitask Model Details}
\label{app:multitask-model}
The multitask model follows the specifications by \cite{davani_dealing_2022}. Specifically, let $BERT$ be the Hugging Face ``bert-base-uncased" model, which takes in a text, $t_i$, and outputs the embedding of the [CLS] token for that text \cite{devlin_bert_2019, noauthor_bert-base-uncased_2023}. Then, let $Lin$ represent a linear layer which takes in the embedding output by $BERT$ and outputs $K$ values, where $K$ is the number of valid annotation classes. We have $M$ of these linear layers, one for each annotator $j$. Finally, let $v_i$ be a single-dimensional array whose $j$th entry is 1 if the corresponding annotation $a_{i, j}$ is not missing (is valid), and 0 if it is missing (is not valid). Finally, let $CE$ represent the cross entropy function of two vectors.

Then, the output of the model $o_i$ for a given text $t_i$ is computed as a single-dimensional array whose $j$th value is $$o_{i, j} = Lin_j(BERT(t_i)).$$ And the loss for the model is computed as $$CE(o_i \odot v_i, a_i).$$

Exact implementation details can be found in our code.

\section{Dataset Details}
\label{app:datasets}
Each dataset consists of two files: a text and annotation file. 
The text file consists of $N$ texts, such that $t_i$ refers to the $i$th text, where $1 \le i \le N$. The annotation file consists of annotations of text, and is a $NxM$ matrix, where $a_{i, j}$ refers to the annotation given by the $j$th annotator for the $i$th text, where $1 \le j \le M$.

For all datasets, $a_{i, j}$ is an integer rating of the text. While different datasets have upper bounds of potential ratings, ratings which are numerically close to one another signify annotations which are semantically close to one another. In other words, for the datasets we use, a rating of 1 is similar to a rating of 2 and less similar to a rating of 5. This is in contrast to standard classification tasks, where class labels may differ significantly in semantics despite being close numerically.

\section{Hyperparameters}
\label{app:hyperparameters}

\subsection{NCF Matrix Factorization}
The hyperparameters for NCF matrix factorization are
\begin{itemize}
    \item Factors: [4, 8, 16, 32, 64, 128]
    \item Learning Rates: [0.001, 0.0005, 0.0001, 0.00005]
\end{itemize}
Hyperparameters are picked automatically through a grid search of all possible values during each run based on whichever hyperparameters achieve the lowest RMSE score on all of the training data.

\subsection{Kernel Matrix Factorization}
The hyperparameters for kernel matrix factorization are:
\begin{itemize}
    \item Factors: [1, 2, 4, 8, 16, 32]
    \item Epochs: [1, 2, 4, 8, 16, 32, 64, 128, 256]
    \item Kernels: [linear, rbf, sigmoid]
    \item Gammas: (always set to auto)
    \item Regularization: [0.1, 0.01, 0.001]
    \item Learning Rate: [0.01, 0.001, 0.0001]
    \item Initial Mean: (always set to 0)
    \item Initial Standard Deviation: (always set to 0.1)
    \item Random Seed: [42 85]
\end{itemize}
The hyperparameters used for each imputation task are picked automatically based on a randomly-chosen held-out validation set consisting of 5\% of the training data.

\subsection{Multitask Model}
The hyperparameters for the Multitask model are:
\begin{itemize}
    \item Epochs: (always set to 10)
    \item Learning rate: (always set to 5e-5)
\end{itemize}

\section{Additional Imputed Prompting Experiments}
\label{app:chatgpt-extra-results}
\begin{table}[h]
	\begin{center}
    \scalebox{0.95}{
	\begin{tabular}{@{}r|ccc@{}}
        \toprule
		Method / Dataset & GHC & SBIC & SChem \\
        \midrule
	  Orig. Dist. & \underline{83.33\%} & \underline{76.67\%} & \textbf{50.00\%} \\
      NCF Dist. & \underline{83.33\%} & \underline{76.67\%} & \textbf{50.00\%} \\
      Maj. Voted & \textbf{88.89\%} & \textbf{83.33\%} & \underline{45.00\%} \\
      \bottomrule
	\end{tabular}
    }
    \end{center}
    \caption{Accuracy of GPT-3 at making individualized predictions for a given text when provided with 1. The original distribution 2. The NCF-imputed distribution and 3. The majority-voted annotation for that text}
     \label{tab:gpt-3}
\end{table}

\begin{table}[h]
    \begin{center}
    \scalebox{0.95}{
        \begin{tabular}{@{}r|ccc@{}}
        \toprule
        Method / Dataset & GHC & SBIC & SChem \\
        \midrule
        Not Imputed & \textbf{0.624} & \textbf{0.653} & \textbf{0.425} \\
        Imputed & 0.471 & 0.594 & 0.312 \\
        \bottomrule
        \end{tabular}
    }
    \end{center}
    \caption{Weighted F1 score of ChatGPT making individualized predictions for one out of three provided individuals. The highest F1 score for each dataset is in \textbf{bold}. Example prompts can be found in Appendix \ref{app:prompts}.}
     \label{tab:gpt-3-annotator-tags}
\end{table}

\begin{table}[h]
    \begin{center}
    \scalebox{1}{
        \begin{tabular}{@{}r|ccc@{}}
        \toprule
        Method / Dataset & GHC & SBIC & SChem \\
        \midrule
        Not Imputed & \textbf{11.020\ms{10.169}} & 10.525\ms{10.484} & \textbf{0.430\ms{0.483}} \\
        Imputed & 12.459\ms{10.241} & \textbf{8.921\ms{9.687}} & 0.533\ms{0.621} \\
        \bottomrule
        \end{tabular}
    }
    \end{center}
    \caption{KL divergence score (and standard deviation) of distributions predicted by ChatGPT compared to the true distributions for the given datasets. Imputation is done via the NCF method. Results are reported from the ``no-context" prompt (see Appendix \ref{app:prompts}).}
     \label{tab:gpt-3-distribution}
\end{table}

In Table \ref{tab:gpt-3} we provide an overview of performance comparing the accuracy of GPT-3 for making individualized predictions when provided with either 1. The original soft label 2. The imputed soft label or 3. The majority-voted label. Note that we expect a lower accuracy for SChem in comparison to SBIC or GHC since SChem has 5 labels, while SBIC and GHC have 3 and 2 labels respectively.

Interestingly, there was \textit{no impact to accuracy} based on whether or not imputed versus original data was used. While Section \ref{sec:results:soft-label-analysis} clearly indicates differences between imputed soft labels and original soft labels, GPT-3 appears to be robust to these differences when making individualized predictions.

We do see that providing the majority-voted annotation rather than the soft label improves performance by roughly 5\% on GHC and 7\% on SBIC. However, it also drops performance on SChem by 5\%. This appears to indicate that for datasets with less labels, providing the majority-voted label enables GPT-3 to make better predictions than if one were to provide a soft label. However, as the number of labels increases, soft labels may provide more informative information for accurate predictions.

In Table \ref{tab:gpt-3-annotator-tags} we display the impact of imputed data on making individualized predictions for one of three annotators whose data was provided in the prompt. The data clearly shows that imputation has a negative impact on ChatGPT's ability to make accurate individualized predictions.

Similar results are shown in Table \ref{tab:gpt-3-distribution} where we display the impact of imputed data on making soft label predictions. A high KL divergence score indicates a worse prediction; for GHC and SChem, imputation seems to harm the predictions, whereas for SBIC, imputation seems to help significantly. However, if we analyze the standard deviation, we see that it is often near if not greater than the mean, indicating a distribution that is skewed highly to the right, and suggesting that any changes in performance are not particularly significant.

\section{Experiments to Support Low-Response Imputation}
\label{app:ablation}
\begin{table*}[t]
  \centering
  \begin{tabular}{|c|ccc|ccc|c|}
    \hline
    \textbf{Version} & \multicolumn{3}{c}{\textbf{Replaced with Original}} \vline & \multicolumn{4}{c}{\textbf{Standard}} \vline \\ \textbf{Prompt}
    & \textbf{``Imputed"} & \textbf{Original} & \textbf{Combined} & \textbf{Imputed} & \textbf{Original} & \textbf{Combined} & \textbf{Low30} \\
    
    \hline
    Politeness & \textbf{0.213} & 0.186 & \textbf{0.199} & 0.094 & 0.186 & 0.085 & 0.14 \\
    \hline
    GHC & \textbf{0.896} & 0.846 & 0.745 & 0.858 & 0.846 & \textbf{0.796} & 0.86 \\
    \hline
    SChem & 0.604 & \textbf{0.359} & 0.563 & \textbf{0.615} & 0.355 & \textbf{0.620} & 0.56 \\
    \hline
    SChem5L & 0.578 & 0.581 & \textbf{0.492} & \textbf{0.595} & 0.581 & 0.407 & \underline{0.65} \\
    \hline
    SBIC & 0.820 & 0.733 & \textbf{0.820} & \textbf{0.831} & 0.733 & 0.760 & \underline{0.95} \\
    \hline
    Sentiment & \textbf{0.513} & 0.063 & 0.513 & 0.496 & \textbf{0.139} & \textbf{0.558} & \underline{0.60} \\
    \hline
    
  \end{tabular}
  \caption{
  Table of F1 scores measuring individualized predictions made by ChatGPT, given data from \textit{high-response-rate} annotators. In the ``Replaced with Original" condition, imputed data is replaced with original data (but the rest of the prompt remains the same). In the ``Standard" condition, imputed data remains imputed. The ``Low30" section copies over the highest F1 score from  from Table \ref{tab:low-response-results}, which uses data from low-response-rate annotators, for direct comparison to these results. Scores are listed in bold if they outcompete their ``Replaced with Original" or ``Standard" counterpart. F1 scores from the ``Low30" column are underlined if they outperform all high-response-rate scores.
  }
  \label{tab:data-ablation}
\end{table*}

\begin{table*}[t]
  \centering
  \begin{tabular}{|c|ccc|ccc|}
    \hline
    \textbf{Version} & \multicolumn{3}{c}{\textbf{Original Prompt}} \vline & \multicolumn{3}{c}{\textbf{Swapped Prompt}} \vline \\ \textbf{Prompt}
    & \textbf{Imputed} & \textbf{Original} & \textbf{Combined} & \textbf{Imputed} & \textbf{Original} & \textbf{Combined} \\
    
    \hline
    Politeness & 0.033 & \underline{0.050} & 0.027 & \textbf{0.067} & \underline{\textbf{0.144}} & \textbf{0.128} \\
    \hline
    GHC &  \underline{0.858} & 0.745 & \textbf{0.846} & \underline{0.858} & \textbf{0.846} & 0.796 \\
    \hline
    SChem & \underline{0.592} & 0.502 & \textbf{0.532} & \underline{0.592} & 0.502 & 0.502 \\
    \hline
    SChem5L & \underline{\textbf{0.747}} & 0.493 & \textbf{0.697} & \underline{0.646} & \textbf{0.519} & 0.600 \\
    \hline
    SBIC & \underline{0.952} & 0.926 & 0.932 & \underline{0.952} & 0.926 & 0.932\\
    \hline
    Sentiment & \underline{0.600} & 0.059 & 0.585 & \textbf{0.604} & \textbf{0.31} & \underline{\textbf{0.638}}\\
    \hline
    
  \end{tabular}
  \caption{F1 Scores of ChatGPT predicting annotations of low-response-rate individuals, but with the text preceding the imputed and original examples swapped. F1 scores are \textbf{bolded} if they are higher than their swapped or original counterpart , and \underline{underlined} if the data (imputed, original, or combined) used in the prompt outperforms other data within the same condition (original or swapped).}
  \label{tab:prompt-ablation}
\end{table*}

Overall, based on the data we have compiled into Table \ref{tab:data-ablation}, there is no clear pattern for annotators with high response rate as to whether using imputed data rather than real data is more beneficial for making individualized predictions. We cannot test if this is the case on the annotators with a low-response-rate, as they do not have enough annotations to replace the imputed annotations. 

Table \ref{tab:prompt-ablation} indicates that swapping the prompts may increase results in some cases, but, again, there\textcolor{red}{\st{'s} is} no clear trend similar to the trend we saw for using imputed data, which can be verified again in this data by noticing that the imputed column consistently outperforms other columns for all datasets but Politeness.

Together, these two experiments show that the increase in F1 score is not due to the text before the prompt, and that it is the moderate increase in examples that imputation can provide, rather than the imputed data itself, that is likely the cause of the increased performance.

\section{Imputed Prompting Prompt Details}
\label{app:prompts}
\subsection{Description of Prompts}
For the highlighted ChatGPT experiment and ablation studies, each of the text portions was chosen from a list of possible options, and each possible combination of these options, along with multiple prompt versions, was used for an initial run on SBIC and politeness. After this initial run, the worst-performing prompts and prompt options were removed, and all datasets were run again. The results reported are the best results among all prompts used. Exact details, including all of the full prompts, prompt options, examples, and outputs can be found in our code.

For GPT-3, we provide either the true (original/non-imputed) soft label, the imputed soft label, or the true majority-voted (aggregate) label for the target text. For the distributional label, we ignore the annotator's actual label when computing the distribution, so as not to cause data leakage. However, when computing the original majority-voted annotation, we leave in the annotator's label for the target example. For the non-highlighted ChatGPT experiments, when making soft label predictions we use the soft label from the imputed data, rather than real data. When making individualized annotation predictions, the example shots are chosen to differ from the original such that the imputed annotation can be used.

\subsection{Prompt Skeletons}
This section displays the skeletons of each of the prompts used. In practice, the portions of the skeleton surrounded by curly braces are replaced with data, which can be seen in Section \ref{app:sec:prompts-full}.

\subsubsection{Highlighted ChatGPT Original Data Skeleton Prompt 1}
\begin{spverbatim}
{dataset_description}

{orig_examples_header}
{orig_examples}

{target_example_header}
{target_example}
{final_words}
\end{spverbatim}

\subsubsection{Highlighted ChatGPT Original Data Skeleton Prompt 2}
\begin{spverbatim}
{dataset_description}

{target_example_header}
{target_example}
{final_words}
\end{spverbatim}

\subsubsection{Highlighted ChatGPT Original Data Skeleton Prompt 3}
\begin{spverbatim}
{dataset_description}

{instructions}
{target_example_header}
{target_example}
{final_words}
\end{spverbatim}

\subsubsection{Highlighted ChatGPT Combined Data Skeleton Prompt}
\begin{spverbatim}
{imputed_examples_header}
{imputed_examples}

{orig_examples_header}
{orig_examples}

{target_example_header}
{target_example}
\end{spverbatim}

\subsubsection{Highlighted ChatGPT Imputed Data Skeleton Prompt 1}
\begin{spverbatim}
{imputed_examples}

{target_example}
\end{spverbatim}

\subsubsection{Highlighted ChatGPT Imputed Data Skeleton Prompt 2}
\begin{spverbatim}
{imputed_examples_header}
{imputed_examples}

{target_example_header}
{target_example}
\end{spverbatim}

\subsubsection{GPT-3 Distributional Skeleton Prompt}
\begin{spverbatim}
Here's a description of a dataset:
{dataset_description}

Given the previous dataset description, your goal is to predict how one of the annotators of the previous dataset would annotate an example from that dataset. You will be given {n_shots} samples of how that particular annotator has responded to other examples and be shown the distributional label of how all annotators have annotated the target example, and will then complete the prediction for the target example as that annotator would.

Here's the samples of how the particular annotator has responded to other examples:
{shots}

Here's how the distributional label of how all annotators have annotated the target example:
{other_shots}

How would the particular annotator annotate the target example?
{target_example_line}
ANSWER:
\end{spverbatim}

\subsubsection{GPT-3 Individual Skeleton Prompt}
\begin{spverbatim}
Here's a description of a dataset:
{dataset_description}

Given the previous dataset description, your goal is to predict how one of the annotators of the previous dataset would annotate an example from that dataset. You will be given {n_shots} samples of how that particular annotator has responded to other examples and {k_shots} sample of how others have annotated the target example, and will then complete the prediction for the target example as that annotator would.

Here's the samples of how the particular annotator has responded to other examples:
{shots}

Here's the samples of how others have annotated the target example:
{other_shots}

How would the particular annotator annotate the target example?
{target_example_line}
ANSWER:
\end{spverbatim}

\subsubsection{GPT-3 Majority-Voted Skeleton Prompt}
\begin{spverbatim}
Here's a description of a dataset:
{dataset_description}

Given the previous dataset description, your goal is to predict how one of the annotators of the previous dataset would annotate an example from that dataset. You will be given {n_shots} samples of how that particular annotator has responded to other examples and be shown what the plurality of annotators gave as a label, and will then complete the prediction for the target example as that annotator would.

Here's the samples of how the particular annotator has responded to other examples:
{shots}

Here's how the plurality of annotators labeled the target example:
{other_shots}

How would the particular annotator annotate the target example?
{target_example_line}
ANSWER:
\end{spverbatim}

\subsubsection{ChatGPT Soft Label Skeleton Prompt}
\begin{spverbatim}
{soft_label_examples}
{prediction_text}
\end{spverbatim}

\subsubsection{(Unused) ChatGPT Contextual Soft Label Skeleton}
\begin{spverbatim}
Here is a description of a dataset:
{dataset_description}

Your goal is to predict the soft label given by the raters on a particular text.

Here are a few examples of texts and their soft label:
{soft_label_examples}

Now, you will make your prediction (if you are unsure, just give your best estimate):
{prediction_text}
\end{spverbatim}

\subsubsection{ChatGPT One of Three Individualized Skeleton Prompt}
\begin{spverbatim}
Description of the dataset:
{dataset_description}

How annotator A has labeled some examples:
{annotator_A_examples}

How annotator B has labeled some examples:
{annotator_B_examples}

How annotator C has labeled some examples:
{annotator_C_examples}

How might annotator {target_annotator} label this example? Output the annotation that seems most likely.
{target_example}
\end{spverbatim}

\subsection{Full Prompts}
\label{app:sec:prompts-full}
This section demonstrates how the skeletons are completed with full data for the prompts that are actually sent to the OpenAI GPT models. Version numbers such as ``v4.-1.0.-1.1" indicate which fillers were used for the prompt. The value ``v4.-1.0.-1.1" indicates that the 5th option was used for the first filler, the second filler was not included in the prompt, the first option was used for the third filler, the second to last filler was not included, and the 2nd option was used for the last filler.

The fillers, in order, are:
\begin{spverbatim}
1. orig_examples_header
2. imputed_examples_header
3. target_example_header
4. instructions
5. final_words
\end{spverbatim}

\vspace{\baselineskip}

As an example, the options for the filler may look like:
\begin{spverbatim}
Fillers for final_words:
1. "Your output should be a single integer corresponding to the label."
2. "Your output should be a single integer and nothing else."
3. "The only valid output is a single integer."
4. "If you output anything other than a single integer, your output will be considered invalid."
5. "If you output anything other than a single integer, your output will harm the integrity of our dataset."
6. "If you output anything other than a single integer (and absolutely nothing else, including explanatory text), your output will invalidate the dataset."
7. "If you output anything other than a single integer (and absolutely nothing else, including explanatory text), your output will invalidate the dataset. So, please only output a single integer."
\end{spverbatim}

Explicit details of all filler options for all experiments can be found in the code.

\subsubsection{Highlighted ChatGPT Original Data Full Prompt 1 (v4.-1.0.-1.1)}
\begin{spverbatim}
In this dataset, people are given rules of thumb, such as "You should be accepting of other people's bodies." People then rate the rule of thumb as to how much of the world population they think would agree with that rule of thumb. 4 means "universal". 3 means "common". 2 means "controversial". 1 means "occasional". 0 means "almost no one".

Examples from the dataset (there may not be any):Examples from the dataset (there may not be any)
Example 1:
Text: It's good to trust those you care about.
Annotation from annotator: 3

Example 2:
Text: It's wrong to grow apart from your partner.
Annotation from annotator: 1

Example 3:
Text: It's good to find proof of cheating you perceive to be happening.
Annotation from annotator: 3

Example 4:
Text: It's good to get a friend's help.
Annotation from annotator: 3

Example 5:
Text: You should always help out your customers when they have a problem.
Annotation from annotator: 3

Example 6:
Text: You shouldn't post inappropriate pictures online.
Annotation from annotator: 3

Example 7:
Text: You shoulld not worry about others being jealous when posting things on facebook.
Annotation from annotator: 3

Example 8:
Text: You should explain to a friend why you cannot give them money if you cannot.
Annotation from annotator: 2

Example 9:
Text: You shouldn't video tape someone without their permission.
Annotation from annotator: 3

Example 10:
Text: It's good to give advice to your friends.
Annotation from annotator: 3

Example 11:
Text: You shouldn't judge people by how they look
Annotation from annotator: 3

Example 12:
Text: It is okay to show emotions.
Annotation from annotator: 2

Example 13:
Text: It's bad to hate babies.
Annotation from annotator: 3

Example 14:
Text: You shouldn't scare your coworkers.
Annotation from annotator: 3

Example 15:
Text: It's not okay to rely on technology for health.
Annotation from annotator: 1

Example 16:
Text: You can't expect love young to last forever.
Annotation from annotator: 3

Example 17:
Text: It's good to want to do the things it takes to start a grown up life.
Annotation from annotator: 3

Example 18:
Text: It's normal for married couples to have sexual relations.
Annotation from annotator: 4

Example 19:
Text: You should always know why you do things.
Annotation from annotator: 3

Example 20:
Text: It's understandable to want to get back at someone.
Annotation from annotator: 3

Example 21:
Text: You shouldn't date a bad partner.
Annotation from annotator: 3

Example 22:
Text: It's ok to make your own reproductive choices.
Annotation from annotator: 2

Example 23:
Text: It's good for a parent to want their children to make the best decisions possible.
Annotation from annotator: 3

Example 24:
Text: It is distracting to pin everyone in an admin team on a discord server.
Annotation from annotator: 3

Example 25:
Text: It is okay to not share something that you did a lonf time ago.
Annotation from annotator: 3

Example 26:
Text: It is good to be yourself.
Annotation from annotator: 3

Example 27:
Text: It's expected that you'll stay home from work when you are sick.
Annotation from annotator: 3

Example 28:
Text: It's inappropriate for a parent to help a teenage boy shower.
Annotation from annotator: 3

Example 29:
Text: It's good to want to be able to back up your words.
Annotation from annotator: 4

Example 30:
Text: You should treat employees well
Annotation from annotator: 3

Target Example:
Text: It's OK not to take life too seriously.
Annotation from annotator:
Your output should be a single integer and nothing else.
\end{spverbatim}

\subsubsection{Highlighted ChatGPT Original Data Full Prompt 2 (v-1.-1.0.-1.0)}
\begin{spverbatim}
In this dataset, people are given rules of thumb, such as "You should be accepting of other people's bodies." People then rate the rule of thumb as to how much of the world population they think would agree with that rule of thumb. 4 means "universal". 3 means "common". 2 means "controversial". 1 means "occasional". 0 means "almost no one".

Target Example:
Text: People shouldn't betray the trust of those they're close to.
Annotation from annotator:
Your output should be a single integer corresponding to the label.
\end{spverbatim}

\subsubsection{Highlighted ChatGPT Original Data Full Prompt 3 (v-1.-1.0.1.0)}
\begin{spverbatim}
In this dataset, people are given rules of thumb, such as "You should be accepting of other people's bodies." People then rate the rule of thumb as to how much of the world population they think would agree with that rule of thumb. 4 means "universal". 3 means "common". 2 means "controversial". 1 means "occasional". 0 means "almost no one".

Now you will make your prediction (if you are unsure, just give your best estimate.) Your output should be an integer label:
Target Example:
Text: People shouldn't betray the trust of those they're close to.
Annotation from annotator:
Your output should be a single integer corresponding to the label.
\end{spverbatim}

\subsubsection{Highlighted ChatGPT Combined Full Prompt (v1.0.0.-1.-1}
\begin{spverbatim}
Estimated Examples:
Example 1:
Text: It's wise to never let yourself get embarrassed by your significant other.
Annotation from annotator: 2

Example 2:
Text: It's not normal to dislike everybody
Annotation from annotator: 3

Example 3:
Text: It's rude to dislike your best friend's boyfriend.
Annotation from annotator: 2

Example 4:
Text: It's okay to be upset if someone you report to at work lets you take the blame for something you didn't do.
Annotation from annotator: 3

Example 5:
Text: You should always respect your partner's choices.
Annotation from annotator: 3

Example 6:
Text: It's wrong to spend other people's money.
Annotation from annotator: 3

Example 7:
Text: You shouldn't be so scared about getting married.
Annotation from annotator: 2

Example 8:
Text: It's wrong to want to hide who you really are.
Annotation from annotator: 2

Example 9:
Text: You shouldn't have sex with people if you have herpes.
Annotation from annotator: 3

Example 10:
Text: You shouldn't video tape someone without their permission.
Annotation from annotator: 3

Example 11:
Text: You should thank someone when they invite tou to thier wedding.
Annotation from annotator: 3

Example 12:
Text: You can't expect your family members to support you if you're lazy.
Annotation from annotator: 3

Example 13:
Text: It's expected that revealing secrets will lead to trouble.
Annotation from annotator: 3

Example 14:
Text: If you are uncomfortable with your partner feeding your kids, you should not let your partner watch your children.
Annotation from annotator: 3

Example 15:
Text: It's good to want to keep your children safe in life.
Annotation from annotator: 3

Example 16:
Text: It's okay to have sex on camera.
Annotation from annotator: 2

Example 17:
Text: It's wrong for partners to lie to each other.
Annotation from annotator: 3

Example 18:
Text: It is reasonable to want to hold onto loving memories after losing someone.
Annotation from annotator: 3

Example 19:
Text: You shouldn't play pranks on people.
Annotation from annotator: 2

Example 20:
Text: It's okay to have an abortion.
Annotation from annotator: 2

Example 21:
Text: It's wrong to think you're responsible for somebody else's life
Annotation from annotator: 2

Example 22:
Text: You should not get angry when your chronically ill relative needs help.
Annotation from annotator: 3

Example 23:
Text: You shouldn't be bitter and salty about petty things.
Annotation from annotator: 3

Example 24:
Text: It's good to play games of chance where you can win money.
Annotation from annotator: 2

Example 25:
Text: Relationships are expected to have clear communication.
Annotation from annotator: 3

Example 26:
Text: You shouldn't try to get sympathy from people just because you're in the hospital.
Annotation from annotator: 2

Example 27:
Text: You shouldn't be friends with someone that talks badly about you behind your back.
Annotation from annotator: 3

Example 28:
Text: It's okay to carry your pet in public as a status symbol
Annotation from annotator: 1

Example 29:
Text: It's not okay to be in a romantic relationship with a teacher.
Annotation from annotator: 3

Example 30:
Text: You can't expect everyone to like you.
Annotation from annotator: 3

Human-rated Examples:
Example 1:
Text: It's good to be honest when you give your kids the sex talk.
Annotation from annotator: 3

Example 2:
Text: It's wrong for people that care about each other to argue.
Annotation from annotator: 2

Example 3:
Text: It's inappropriate to make passes at an underling.
Annotation from annotator: 3

Example 4:
Text: Couple should work together to solve problems
Annotation from annotator: 3

Example 5:
Text: It's okay to throw away compostables.
Annotation from annotator: 2

Example 6:
Text: It's bad to smoke.
Annotation from annotator: 3

Example 7:
Text: You can't expect someone to pay off their loans immediately.
Annotation from annotator: 3

Example 8:
Text: It's not necessary to be strong
Annotation from annotator: 3

Example 9:
Text: You shouldn't put up an antenna if it's snowing outside.
Annotation from annotator: 1

Example 10:
Text: It's good to have a best friend.
Annotation from annotator: 3

Example 11:
Text: It's bad to lose track of your belongings.
Annotation from annotator: 3

Example 12:
Text: It's not okay to have to rely on special tools when they aren't really needed.
Annotation from annotator: 2

Target Example:
Text: People shouldn't betray the trust of those they're close to.
Annotation from annotator:
\end{spverbatim}

\subsubsection{Highlighted ChatGPT Imputed Data Full Prompt 1 (v-1.-1.-1.-1.-1)}
\begin{spverbatim}
Example 1:
Text: It's wise to never let yourself get embarrassed by your significant other.
Annotation from annotator: 2

Example 2:
Text: It's not normal to dislike everybody
Annotation from annotator: 3

Example 3:
Text: It's rude to dislike your best friend's boyfriend.
Annotation from annotator: 2

Example 4:
Text: It's okay to be upset if someone you report to at work lets you take the blame for something you didn't do.
Annotation from annotator: 3

Example 5:
Text: You should always respect your partner's choices.
Annotation from annotator: 3

Example 6:
Text: It's wrong to spend other people's money.
Annotation from annotator: 3

Example 7:
Text: You shouldn't be so scared about getting married.
Annotation from annotator: 2

Example 8:
Text: It's wrong to want to hide who you really are.
Annotation from annotator: 2

Example 9:
Text: You shouldn't have sex with people if you have herpes.
Annotation from annotator: 3

Example 10:
Text: You shouldn't video tape someone without their permission.
Annotation from annotator: 3

Example 11:
Text: You should thank someone when they invite tou to thier wedding.
Annotation from annotator: 3

Example 12:
Text: You can't expect your family members to support you if you're lazy.
Annotation from annotator: 3

Example 13:
Text: It's expected that revealing secrets will lead to trouble.
Annotation from annotator: 3

Example 14:
Text: If you are uncomfortable with your partner feeding your kids, you should not let your partner watch your children.
Annotation from annotator: 3

Example 15:
Text: It's good to want to keep your children safe in life.
Annotation from annotator: 3

Example 16:
Text: It's okay to have sex on camera.
Annotation from annotator: 2

Example 17:
Text: It's wrong for partners to lie to each other.
Annotation from annotator: 3

Example 18:
Text: It is reasonable to want to hold onto loving memories after losing someone.
Annotation from annotator: 3

Example 19:
Text: You shouldn't play pranks on people.
Annotation from annotator: 2

Example 20:
Text: It's okay to have an abortion.
Annotation from annotator: 2

Example 21:
Text: It's wrong to think you're responsible for somebody else's life
Annotation from annotator: 2

Example 22:
Text: You should not get angry when your chronically ill relative needs help.
Annotation from annotator: 3

Example 23:
Text: You shouldn't be bitter and salty about petty things.
Annotation from annotator: 3

Example 24:
Text: It's good to play games of chance where you can win money.
Annotation from annotator: 2

Example 25:
Text: Relationships are expected to have clear communication.
Annotation from annotator: 3

Example 26:
Text: You shouldn't try to get sympathy from people just because you're in the hospital.
Annotation from annotator: 2

Example 27:
Text: You shouldn't be friends with someone that talks badly about you behind your back.
Annotation from annotator: 3

Example 28:
Text: It's okay to carry your pet in public as a status symbol
Annotation from annotator: 1

Example 29:
Text: It's not okay to be in a romantic relationship with a teacher.
Annotation from annotator: 3

Example 30:
Text: You can't expect everyone to like you.
Annotation from annotator: 3

Text: People shouldn't betray the trust of those they're close to.
Annotation from annotator:
\end{spverbatim}

\subsubsection{Highlighted ChatGPT Imputed Data Full Prompt 2 (v-1.0.0.-1.-1)}
\begin{spverbatim}
Estimated Examples:
Example 1:
Text: It's wise to never let yourself get embarrassed by your significant other.
Annotation from annotator: 2

Example 2:
Text: It's not normal to dislike everybody
Annotation from annotator: 3

Example 3:
Text: It's rude to dislike your best friend's boyfriend.
Annotation from annotator: 2

Example 4:
Text: It's okay to be upset if someone you report to at work lets you take the blame for something you didn't do.
Annotation from annotator: 3

Example 5:
Text: You should always respect your partner's choices.
Annotation from annotator: 3

Example 6:
Text: It's wrong to spend other people's money.
Annotation from annotator: 3

Example 7:
Text: You shouldn't be so scared about getting married.
Annotation from annotator: 2

Example 8:
Text: It's wrong to want to hide who you really are.
Annotation from annotator: 2

Example 9:
Text: You shouldn't have sex with people if you have herpes.
Annotation from annotator: 3

Example 10:
Text: You shouldn't video tape someone without their permission.
Annotation from annotator: 3

Example 11:
Text: You should thank someone when they invite tou to thier wedding.
Annotation from annotator: 3

Example 12:
Text: You can't expect your family members to support you if you're lazy.
Annotation from annotator: 3

Example 13:
Text: It's expected that revealing secrets will lead to trouble.
Annotation from annotator: 3

Example 14:
Text: If you are uncomfortable with your partner feeding your kids, you should not let your partner watch your children.
Annotation from annotator: 3

Example 15:
Text: It's good to want to keep your children safe in life.
Annotation from annotator: 3

Example 16:
Text: It's okay to have sex on camera.
Annotation from annotator: 2

Example 17:
Text: It's wrong for partners to lie to each other.
Annotation from annotator: 3

Example 18:
Text: It is reasonable to want to hold onto loving memories after losing someone.
Annotation from annotator: 3

Example 19:
Text: You shouldn't play pranks on people.
Annotation from annotator: 2

Example 20:
Text: It's okay to have an abortion.
Annotation from annotator: 2

Example 21:
Text: It's wrong to think you're responsible for somebody else's life
Annotation from annotator: 2

Example 22:
Text: You should not get angry when your chronically ill relative needs help.
Annotation from annotator: 3

Example 23:
Text: You shouldn't be bitter and salty about petty things.
Annotation from annotator: 3

Example 24:
Text: It's good to play games of chance where you can win money.
Annotation from annotator: 2

Example 25:
Text: Relationships are expected to have clear communication.
Annotation from annotator: 3

Example 26:
Text: You shouldn't try to get sympathy from people just because you're in the hospital.
Annotation from annotator: 2

Example 27:
Text: You shouldn't be friends with someone that talks badly about you behind your back.
Annotation from annotator: 3

Example 28:
Text: It's okay to carry your pet in public as a status symbol
Annotation from annotator: 1

Example 29:
Text: It's not okay to be in a romantic relationship with a teacher.
Annotation from annotator: 3

Example 30:
Text: You can't expect everyone to like you.
Annotation from annotator: 3

Target Example:
Text: People shouldn't betray the trust of those they're close to.
Annotation from annotator:
\end{spverbatim}

\subsubsection{GPT-3 Non-imputed Distributional Full Prompt}
\begin{spverbatim}
Here's a description of a dataset:
In this dataset, people are given rules of thumb, such as "You should be accepting of other people's bodies." People then rate the rule of thumb as to how much of the world population they think would agree with that rule of thumb. 4 means "universal". 3 means "common". 2 means "controversial". 1 means "occasional". 0 means "almost no one".

Given the previous dataset description, your goal is to predict how one of the annotators of the previous dataset would annotate an example from that dataset. You will be given 4 samples of how that particular annotator has responded to other examples and be shown the distributional label of how all annotators have annotated the target example, and will then complete the prediction for the target example as that annotator would.

Here's the samples of how the particular annotator has responded to other examples:
1.
EXAMPLE: It's good to be honest when you give your kids the sex talk.
ANSWER: 4

2.
EXAMPLE: It's good to want others to celebrate your joys with you.
ANSWER: 3

3.
EXAMPLE: It is dangerous for a doctor to misdiagnose a patient.
ANSWER: 4

4.
EXAMPLE: It's bad for a child to be in danger.
ANSWER: 4

Here's how the distributional label of how all annotators have annotated the target example:
0.00% of people responded with 0
18.00% of people responded with 1
76.00% of people responded with 2
6.00% of people responded with 3

How would the particular annotator annotate the target example?
EXAMPLE: It's expected that families will not get along well with their in-laws.
ANSWER:
\end{spverbatim}

\subsubsection{GPT-3 Imputed Distributional Full Prompt}
\begin{spverbatim}
Here's a description of a dataset:
In this dataset, people are given rules of thumb, such as "You should be accepting of other people's bodies." People then rate the rule of thumb as to how much of the world population they think would agree with that rule of thumb. 4 means "universal". 3 means "common". 2 means "controversial". 1 means "occasional". 0 means "almost no one".

Given the previous dataset description, your goal is to predict how one of the annotators of the previous dataset would annotate an example from that dataset. You will be given 4 samples of how that particular annotator has responded to other examples and be shown the distributional label of how all annotators have annotated the target example, and will then complete the prediction for the target example as that annotator would.

Here's the samples of how the particular annotator has responded to other examples:
1.
EXAMPLE: It's good to be honest when you give your kids the sex talk.
ANSWER: 4

2.
EXAMPLE: It's good to want others to celebrate your joys with you.
ANSWER: 3

3.
EXAMPLE: It is dangerous for a doctor to misdiagnose a patient.
ANSWER: 4

4.
EXAMPLE: It's bad for a child to be in danger.
ANSWER: 4

Here's how the distributional label of how all annotators have annotated the target example:
0.00% of people responded with 0
10.00% of people responded with 1
85.00% of people responded with 2
5.00% of people responded with 3

How would the particular annotator annotate the target example?
EXAMPLE: It's expected that families will not get along well with their in-laws.
ANSWER:
\end{spverbatim}

\subsubsection{GPT-3 Individual Full Prompt}
\begin{spverbatim}
Here's a description of a dataset:
In this dataset, people are given rules of thumb, such as "You should be accepting of other people's bodies." People then rate the rule of thumb as to how much of the world population they think would agree with that rule of thumb. 4 means "universal". 3 means "common". 2 means "controversial". 1 means "occasional". 0 means "almost no one".

Given the previous dataset description, your goal is to predict how one of the annotators of the previous dataset would annotate an example from that dataset. You will be given 4 samples of how that particular annotator has responded to other examples and 49 sample of how others have annotated the target example, and will then complete the prediction for the target example as that annotator would.

Here's the samples of how the particular annotator has responded to other examples:
1.
EXAMPLE: It's good to be honest when you give your kids the sex talk.
ANSWER: 4

2.
EXAMPLE: It's good to want others to celebrate your joys with you.
ANSWER: 3

3.
EXAMPLE: It is dangerous for a doctor to misdiagnose a patient.
ANSWER: 4

4.
EXAMPLE: It's bad for a child to be in danger.
ANSWER: 4

Here's the samples of how others have annotated the target example:
1.
EXAMPLE: It's expected that families will not get along well with their in-laws.
ANSWER: 1

2.
EXAMPLE: It's expected that families will not get along well with their in-laws.
ANSWER: 2

3.
EXAMPLE: It's expected that families will not get along well with their in-laws.
ANSWER: 2

4.
EXAMPLE: It's expected that families will not get along well with their in-laws.
ANSWER: 2

5.
EXAMPLE: It's expected that families will not get along well with their in-laws.
ANSWER: 2

6.
EXAMPLE: It's expected that families will not get along well with their in-laws.
ANSWER: 2

7.
EXAMPLE: It's expected that families will not get along well with their in-laws.
ANSWER: 3

8.
EXAMPLE: It's expected that families will not get along well with their in-laws.
ANSWER: 1

9.
EXAMPLE: It's expected that families will not get along well with their in-laws.
ANSWER: 3

10.
EXAMPLE: It's expected that families will not get along well with their in-laws.
ANSWER: 1

11.
EXAMPLE: It's expected that families will not get along well with their in-laws.
ANSWER: 1

12.
EXAMPLE: It's expected that families will not get along well with their in-laws.
ANSWER: 2

13.
EXAMPLE: It's expected that families will not get along well with their in-laws.
ANSWER: 2

14.
EXAMPLE: It's expected that families will not get along well with their in-laws.
ANSWER: 2

15.
EXAMPLE: It's expected that families will not get along well with their in-laws.
ANSWER: 2

16.
EXAMPLE: It's expected that families will not get along well with their in-laws.
ANSWER: 1

17.
EXAMPLE: It's expected that families will not get along well with their in-laws.
ANSWER: 2

18.
EXAMPLE: It's expected that families will not get along well with their in-laws.
ANSWER: 2

19.
EXAMPLE: It's expected that families will not get along well with their in-laws.
ANSWER: 2

20.
EXAMPLE: It's expected that families will not get along well with their in-laws.
ANSWER: 2

21.
EXAMPLE: It's expected that families will not get along well with their in-laws.
ANSWER: 2

22.
EXAMPLE: It's expected that families will not get along well with their in-laws.
ANSWER: 2

23.
EXAMPLE: It's expected that families will not get along well with their in-laws.
ANSWER: 2

24.
EXAMPLE: It's expected that families will not get along well with their in-laws.
ANSWER: 2

25.
EXAMPLE: It's expected that families will not get along well with their in-laws.
ANSWER: 2

26.
EXAMPLE: It's expected that families will not get along well with their in-laws.
ANSWER: 2

27.
EXAMPLE: It's expected that families will not get along well with their in-laws.
ANSWER: 2

28.
EXAMPLE: It's expected that families will not get along well with their in-laws.
ANSWER: 2

29.
EXAMPLE: It's expected that families will not get along well with their in-laws.
ANSWER: 2

30.
EXAMPLE: It's expected that families will not get along well with their in-laws.
ANSWER: 2

31.
EXAMPLE: It's expected that families will not get along well with their in-laws.
ANSWER: 2

32.
EXAMPLE: It's expected that families will not get along well with their in-laws.
ANSWER: 3

33.
EXAMPLE: It's expected that families will not get along well with their in-laws.
ANSWER: 2

34.
EXAMPLE: It's expected that families will not get along well with their in-laws.
ANSWER: 2

35.
EXAMPLE: It's expected that families will not get along well with their in-laws.
ANSWER: 2

36.
EXAMPLE: It's expected that families will not get along well with their in-laws.
ANSWER: 2

37.
EXAMPLE: It's expected that families will not get along well with their in-laws.
ANSWER: 2

38.
EXAMPLE: It's expected that families will not get along well with their in-laws.
ANSWER: 2

39.
EXAMPLE: It's expected that families will not get along well with their in-laws.
ANSWER: 2

40.
EXAMPLE: It's expected that families will not get along well with their in-laws.
ANSWER: 2

41.
EXAMPLE: It's expected that families will not get along well with their in-laws.
ANSWER: 1

42.
EXAMPLE: It's expected that families will not get along well with their in-laws.
ANSWER: 2

43.
EXAMPLE: It's expected that families will not get along well with their in-laws.
ANSWER: 2

44.
EXAMPLE: It's expected that families will not get along well with their in-laws.
ANSWER: 1

45.
EXAMPLE: It's expected that families will not get along well with their in-laws.
ANSWER: 2

46.
EXAMPLE: It's expected that families will not get along well with their in-laws.
ANSWER: 2

47.
EXAMPLE: It's expected that families will not get along well with their in-laws.
ANSWER: 2

48.
EXAMPLE: It's expected that families will not get along well with their in-laws.
ANSWER: 1

49.
EXAMPLE: It's expected that families will not get along well with their in-laws.
ANSWER: 2

How would the particular annotator annotate the target example?
EXAMPLE: It's expected that families will not get along well with their in-laws.
ANSWER:
\end{spverbatim}

\subsubsection{GPT-3 Majority-Voted Full Prompt}
\begin{spverbatim}    
Here's a description of a dataset:
In this dataset, people are given rules of thumb, such as "You should be accepting of other people's bodies." People then rate the rule of thumb as to how much of the world population they think would agree with that rule of thumb. 4 means "universal". 3 means "common". 2 means "controversial". 1 means "occasional". 0 means "almost no one".

Given the previous dataset description, your goal is to predict how one of the annotators of the previous dataset would annotate an example from that dataset. You will be given 4 samples of how that particular annotator has responded to other examples and be shown what the plurality of annotators gave as a label, and will then complete the prediction for the target example as that annotator would.

Here's the samples of how the particular annotator has responded to other examples:
1.
EXAMPLE: It's good to be honest when you give your kids the sex talk.
ANSWER: 4

2.
EXAMPLE: It's good to want others to celebrate your joys with you.
ANSWER: 3

3.
EXAMPLE: It is dangerous for a doctor to misdiagnose a patient.
ANSWER: 4

4.
EXAMPLE: It's bad for a child to be in danger.
ANSWER: 4

Here's how the plurality of annotators labeled the target example:
1.
EXAMPLE: It's expected that families will not get along well with their in-laws.
ANSWER: 2

How would the particular annotator annotate the target example?
EXAMPLE: It's expected that families will not get along well with their in-laws.
ANSWER:
\end{spverbatim}

\subsubsection{ChatGPT Non-imputed Soft Label Full Prompt}
\begin{spverbatim}
Example 390
Text: It's hurtful for  a parent to enroll a child in classes without permission.
Soft labels:
0.00% of responsive annotators labeled the text with 0
14.00% of responsive annotators labeled the text with 1
54.00% of responsive annotators labeled the text with 2
28.00% of responsive annotators labeled the text with 3
4.00% of responsive annotators labeled the text with 4

Example 261
Text: It's okay to enjoy powdered milk.
Soft labels:
0.00% of responsive annotators labeled the text with 0
2.00% of responsive annotators labeled the text with 1
16.00% of responsive annotators labeled the text with 2
52.00% of responsive annotators labeled the text with 3
30.00% of responsive annotators labeled the text with 4

Example 20
Text: It's good to find proof of cheating you perceive to be happening.
Soft labels:
0.00% of responsive annotators labeled the text with 0
0.00% of responsive annotators labeled the text with 1
16.00% of responsive annotators labeled the text with 2
64.00% of responsive annotators labeled the text with 3
20.00% of responsive annotators labeled the text with 4

Example 84
Text: People shouldn't betray the trust of those they're close to.
Soft labels:
0.00% of responsive annotators labeled the text with 0
0.00% of responsive annotators labeled the text with 1
0.00% of responsive annotators labeled the text with 2
44.00% of responsive annotators labeled the text with 3
56.00% of responsive annotators labeled the text with 4
Target Text: You should always question someone's character when they say something inappropriate.
Soft labels:
\end{spverbatim}

\subsubsection{ChatGPT Imputed Soft Label Full Prompt}
\begin{spverbatim}
Example 390
Text: It's hurtful for  a parent to enroll a child in classes without permission.
Soft labels:
0.00% of responsive annotators labeled the text with 0
7.00% of responsive annotators labeled the text with 1
67.00% of responsive annotators labeled the text with 2
24.00% of responsive annotators labeled the text with 3
2.00% of responsive annotators labeled the text with 4

Example 261
Text: It's okay to enjoy powdered milk.
Soft labels:
0.00% of responsive annotators labeled the text with 0
1.00% of responsive annotators labeled the text with 1
8.00% of responsive annotators labeled the text with 2
64.00% of responsive annotators labeled the text with 3
27.00% of responsive annotators labeled the text with 4

Example 20
Text: It's good to find proof of cheating you perceive to be happening.
Soft labels:
0.00% of responsive annotators labeled the text with 0
0.00% of responsive annotators labeled the text with 1
8.00% of responsive annotators labeled the text with 2
78.00% of responsive annotators labeled the text with 3
14.00% of responsive annotators labeled the text with 4

Example 84
Text: People shouldn't betray the trust of those they're close to.
Soft labels:
0.00% of responsive annotators labeled the text with 0
0.00% of responsive annotators labeled the text with 1
0.00% of responsive annotators labeled the text with 2
47.00% of responsive annotators labeled the text with 3
53.00% of responsive annotators labeled the text with 4
Target Text: You should always question someone's character when they say something inappropriate.
Soft labels:
\end{spverbatim}

\subsubsection{ChatGPT Non-imputed One of Three Full Prompt}
\begin{spverbatim}
Description of the dataset:
In this dataset, people are given rules of thumb, such as "You should be accepting of other people's bodies." People then rate the rule of thumb as to how much of the world population they think would agree with that rule of thumb. 4 means "universal". 3 means "common". 2 means "controversial". 1 means "occasional". 0 means "almost no one".

How annotator A has labeled some examples:
Example 1:
Text: You should be focused on learning while at school.
Annotation from annotator: 3.0

Example 2:
Text: It is dangerous for a doctor to misdiagnose a patient.
Annotation from annotator: 3.0

Example 3:
Text: It's understandable if you can't keep your political leanings out of your religion lectures.
Annotation from annotator: 3.0

Example 4:
Text: You shouldn't date a bad partner.
Annotation from annotator: 3.0

How annotator B has labeled some examples:
Example 1:
Text: You shouldn't be with someone who sleeps with your kids.
Annotation from annotator: 4.0

Example 2:
Text: You shouldn't have sex with people if you have herpes.
Annotation from annotator: 3.0

Example 3:
Text: You shouldn't be crazy about things.
Annotation from annotator: 2.0

Example 4:
Text: If you don't want to crash, you shouldn't drive recklessly.
Annotation from annotator: 3.0

How annotator C has labeled some examples:
Example 1:
Text: You should only call the cops for emergencies.
Annotation from annotator: 3.0

Example 2:
Text: It is dangerous for a grandparent to risk a fall.
Annotation from annotator: 3.0

Example 3:
Text: Employers are expected to let employees go when business is down.
Annotation from annotator: 4.0

Example 4:
Text: People shouldn't assume it's okay to take something just because it's not nailed down.
Annotation from annotator: 4.0

How might annotator A label this example? Output the annotation that seems most likely.
Text: You should always question someone's character when they say something inappropriate.
Annotation from annotator:
\end{spverbatim}

\subsubsection{ChatGPT Imputed One of Three Full Prompt}
\begin{spverbatim}
Description of the dataset:
In this dataset, people are given rules of thumb, such as "You should be accepting of other people's bodies." People then rate the rule of thumb as to how much of the world population they think would agree with that rule of thumb. 4 means "universal". 3 means "common". 2 means "controversial". 1 means "occasional". 0 means "almost no one".

How annotator A has labeled some examples:
Example 1:
Text: It's hurtful for  a parent to enroll a child in classes without permission.
Annotation from annotator: 2.0

Example 2:
Text: It's okay to enjoy powdered milk.
Annotation from annotator: 3.0

Example 3:
Text: It's good to find proof of cheating you perceive to be happening.
Annotation from annotator: 3.0

Example 4:
Text: People shouldn't betray the trust of those they're close to.
Annotation from annotator: 3.0

How annotator B has labeled some examples:
Example 1:
Text: It's hurtful for  a parent to enroll a child in classes without permission.
Annotation from annotator: 2.0

Example 2:
Text: It's okay to enjoy powdered milk.
Annotation from annotator: 3.0

Example 3:
Text: It's good to find proof of cheating you perceive to be happening.
Annotation from annotator: 3.0

Example 4:
Text: People shouldn't betray the trust of those they're close to.
Annotation from annotator: 3.0

How annotator C has labeled some examples:
Example 1:
Text: It's hurtful for  a parent to enroll a child in classes without permission.
Annotation from annotator: 2.0

Example 2:
Text: It's okay to enjoy powdered milk.
Annotation from annotator: 2.0

Example 3:
Text: It's good to find proof of cheating you perceive to be happening.
Annotation from annotator: 3.0

Example 4:
Text: People shouldn't betray the trust of those they're close to.
Annotation from annotator: 4.0

How might annotator A label this example? Output the annotation that seems most likely.
Text: You should always question someone's character when they say something inappropriate.
Annotation from annotator:
\end{spverbatim}

\section{PCA Results}
\label{app:pca}
\begin{figure*}[t]
\centering

\begin{subfigure}
\centering
\includegraphics[width=0.95\textwidth, clip]{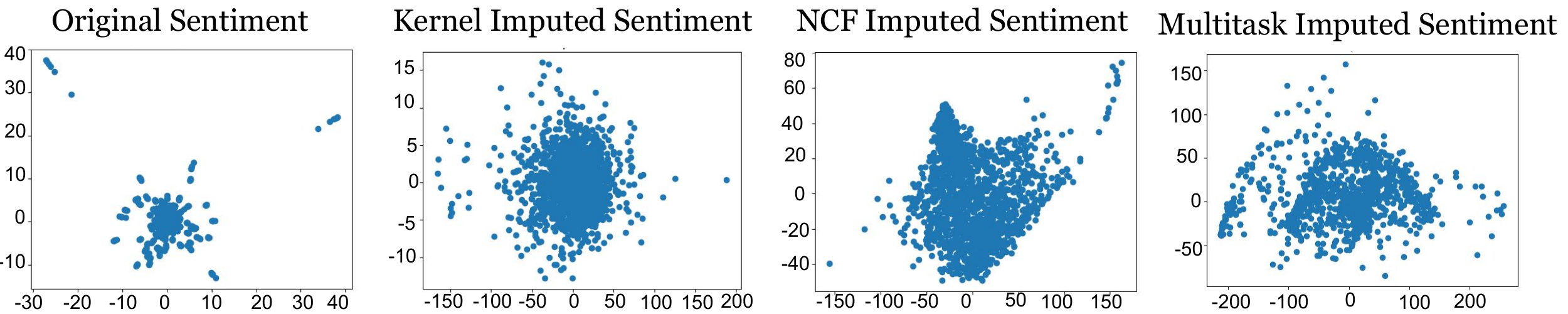}
\end{subfigure}

\begin{subfigure}
\centering
\includegraphics[width=0.95\textwidth,clip]{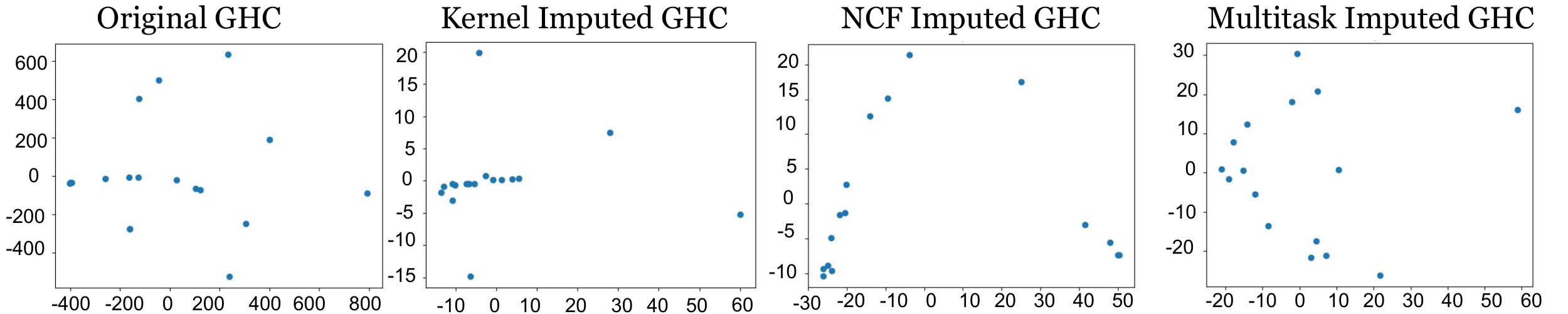}
\end{subfigure}

\begin{subfigure}
\centering
\includegraphics[width=0.95\textwidth,clip]{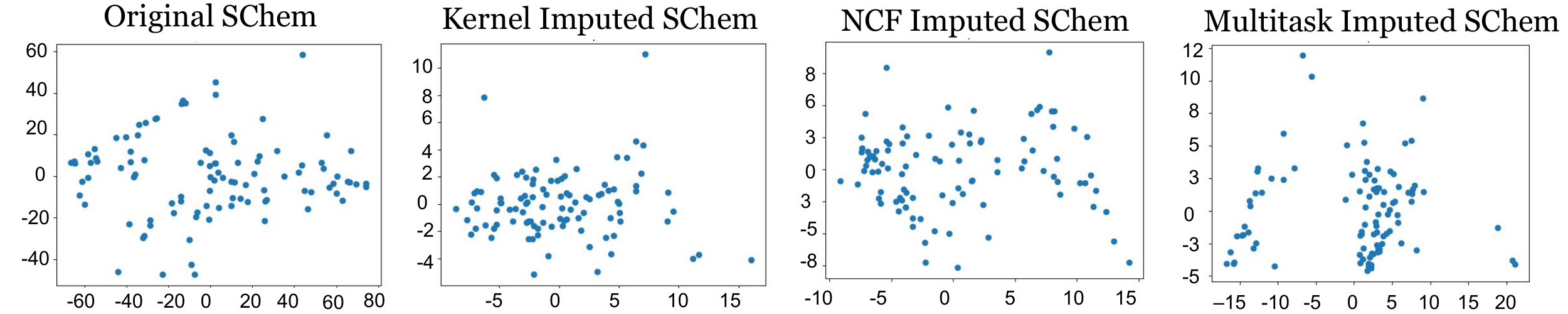}
\end{subfigure}

\begin{subfigure}
\centering
\includegraphics[width=0.95\textwidth,clip] {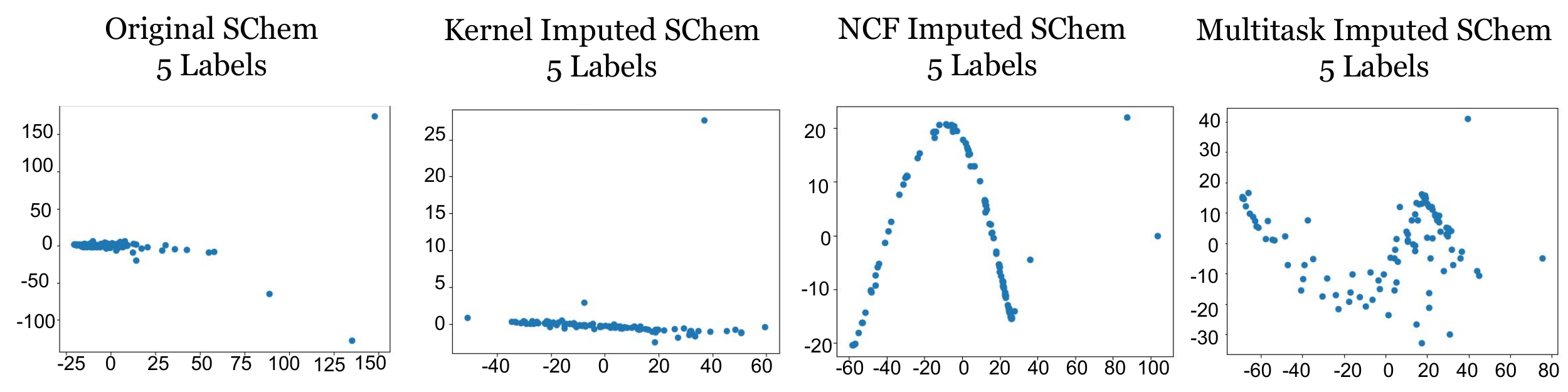}
\end{subfigure}

\begin{subfigure}
\centering
\includegraphics[width=0.95\textwidth, clip]{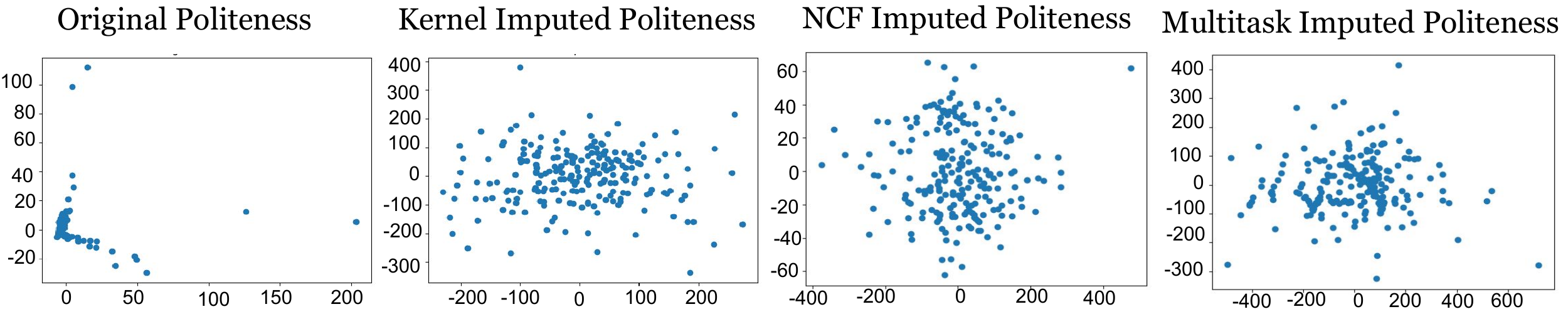}
\end{subfigure}

\caption{PCA projections of each of the datasets after different forms of imputation.}
\label{fig:all-pca}
\end{figure*}

One of the fundamental aspects of imputation methods is how they treat and interpret data. In the 2-dimensional scatter plot based on the first two principal components of imputed datasets, clear variations can be observed across different imputation techniques. This visualization underscores the unique characteristics of each imputation method. Refer to Figure \ref{fig:all-pca} for a detailed comparison.

\section{Variance and Disagreement}
\label{app:variance-and-disagreement}
\begin{figure*}[t]
\centering

\begin{subfigure}
\centering
\includegraphics[width=0.45\textwidth, clip]{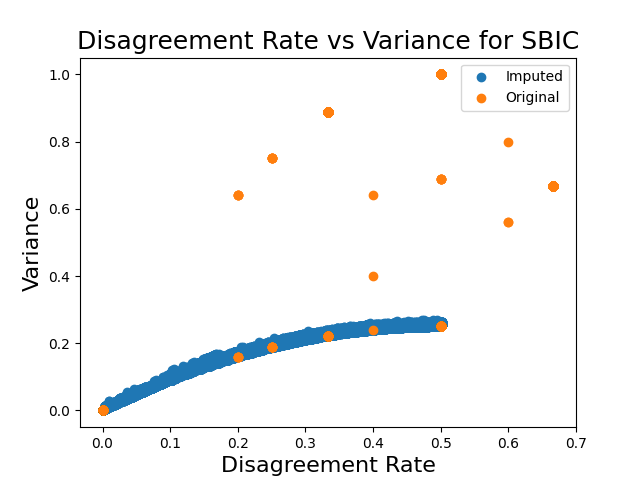}
\includegraphics[width=0.45\textwidth, clip]{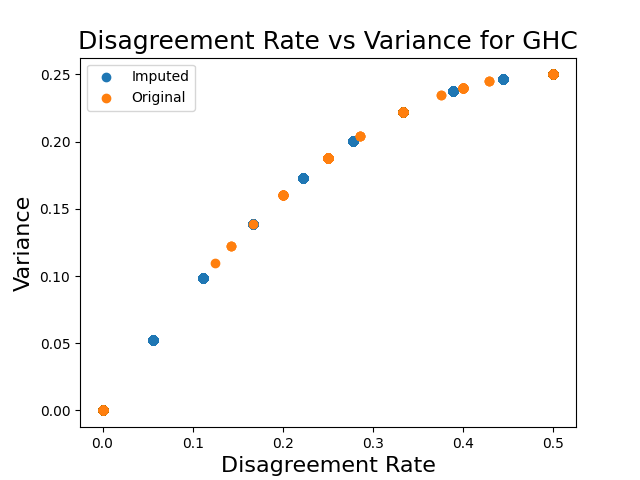}
\end{subfigure}

\begin{subfigure}
\centering
\includegraphics[width=0.45\textwidth, clip]{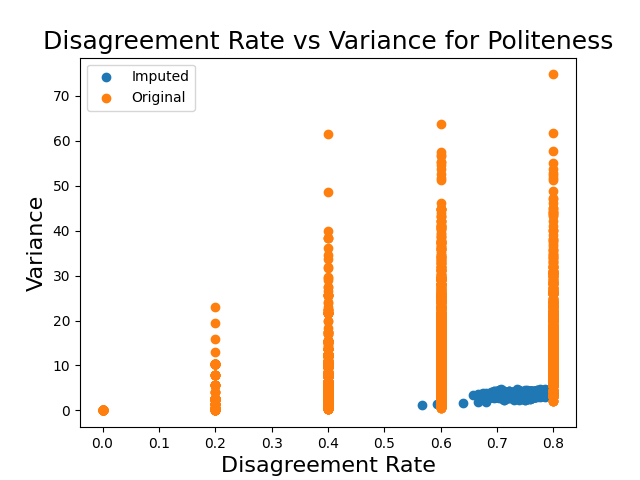}
\includegraphics[width=0.45\textwidth, clip]{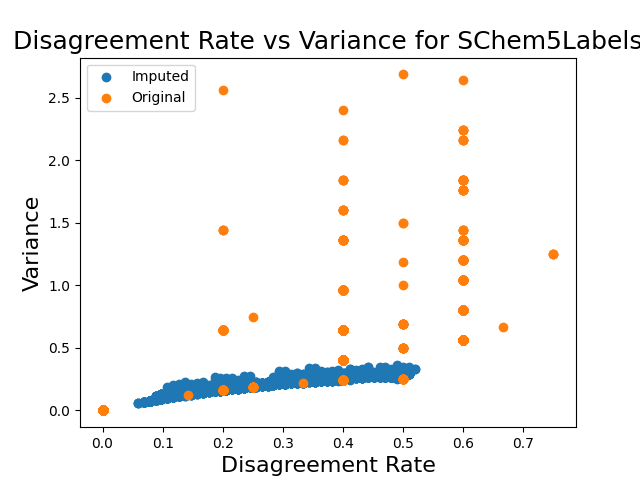}
\end{subfigure}

\begin{subfigure}
\centering
\includegraphics[width=0.45\textwidth, clip]{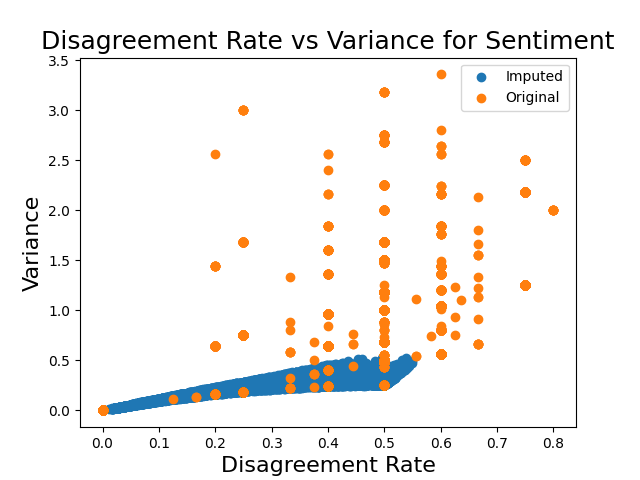}
\end{subfigure}

\caption{Visualizations of the decrease in variance after imputation. Orange is the original dataset, while blue is the imputed data. Each data point represents an example in the dataset. Variance is the variance among annotations for that example, and disagreement rate is the percentage of annotations that disagreed with the majority annotation. Vertical lines in the original dataset data appear because there are only a few annotators for most examples in the original datasets, meaning that the disagreement can only take on a few particular values.}
\label{fig:all-disagreement}
\end{figure*}

Post-imputation, a notable observation is the drop in variance, as shown in Figure \ref{fig:all-disagreement}. This phenomenon can be attributed to the fact that most imputation methods tend to approximate missing values based on observed patterns in the data, leading to a convergence of values around certain estimates. 

\section{Soft Label Analysis Extra Examples}
\label{app:disagreement-website}
\begin{figure*}[h]
    \centering
    \includegraphics[width=0.99\linewidth]{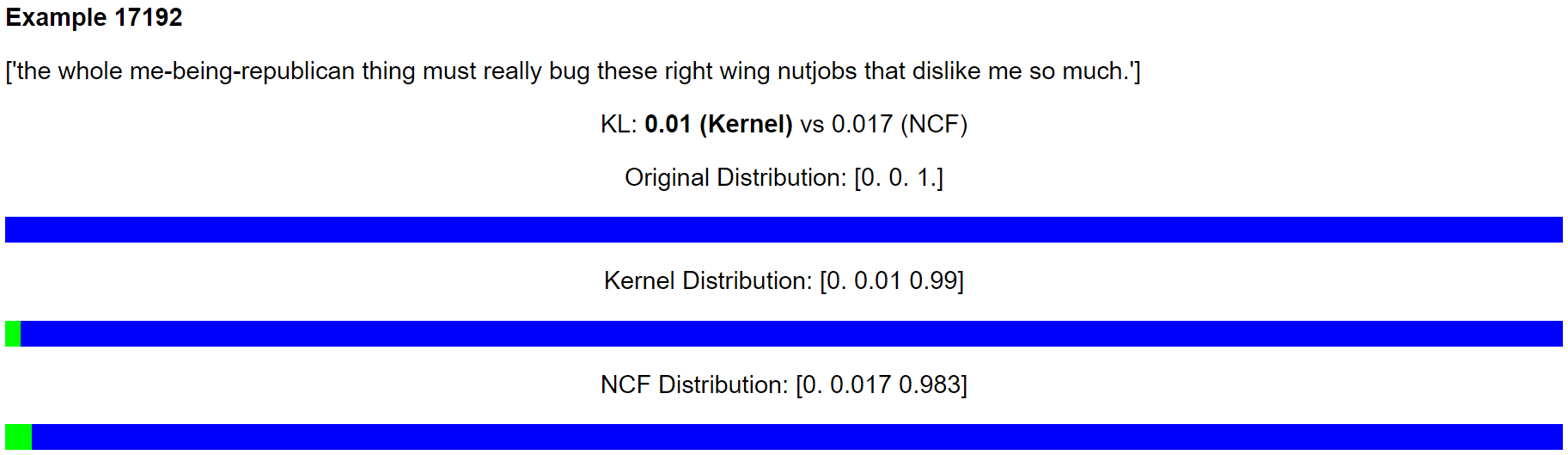}
    \includegraphics[width=0.99\linewidth]{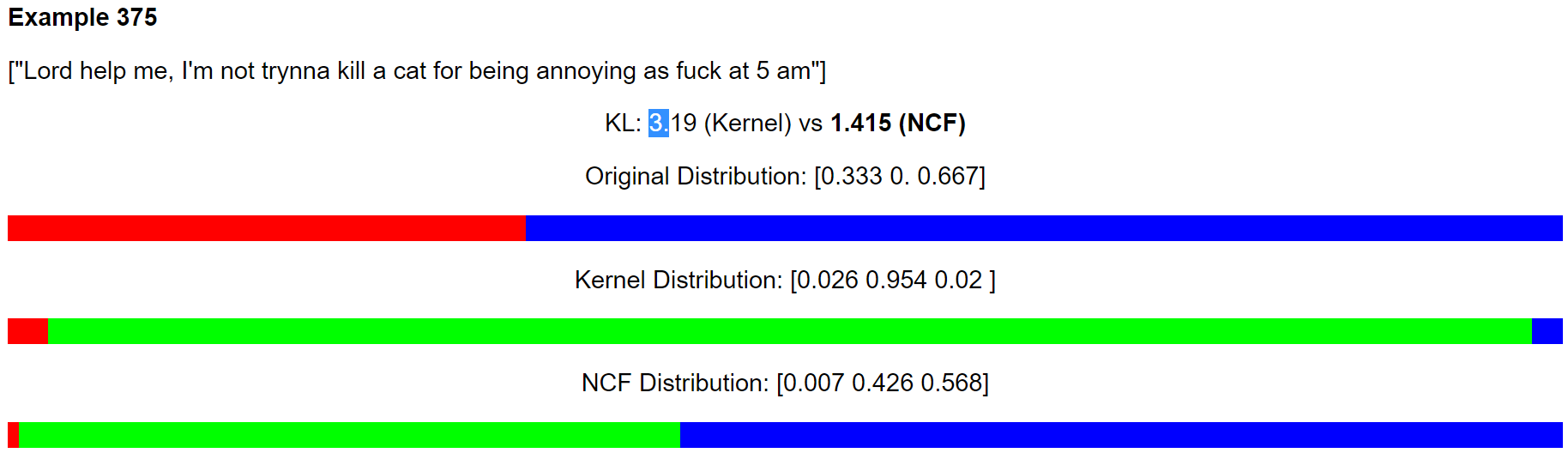}
    \includegraphics[width=0.99\linewidth]{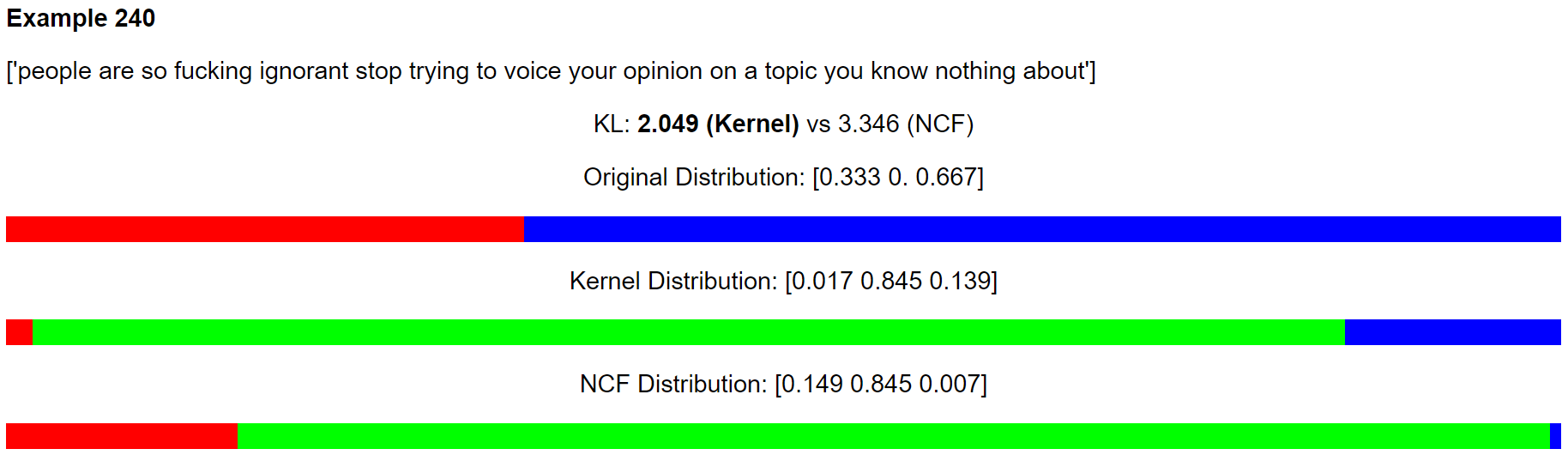}
    \includegraphics[width=0.99\linewidth]{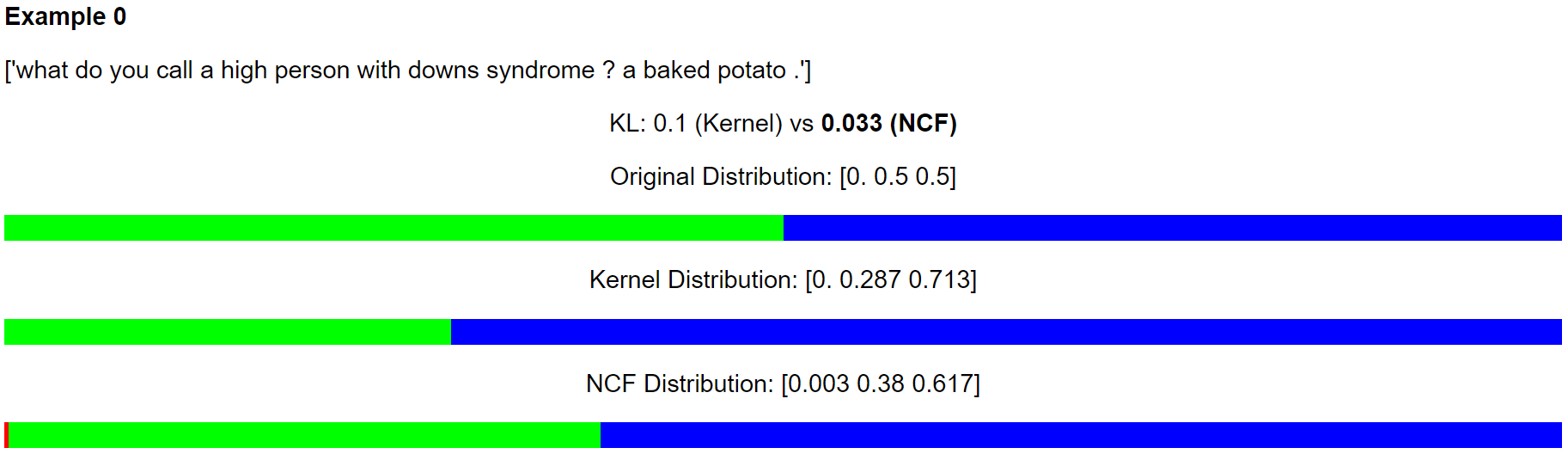}
    \caption{Examples from the SBIC dataset.}
    \label{fig:SBIC-website}
\end{figure*}

\begin{figure*}[t]
    \centering
    \includegraphics[width=0.99\linewidth]{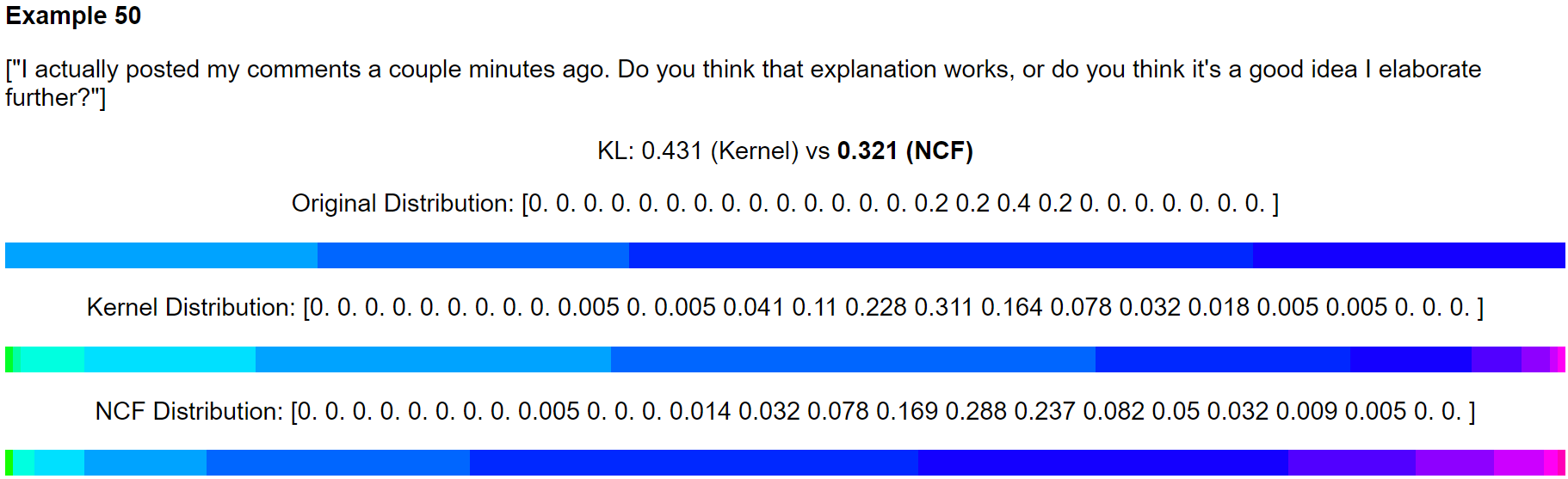}
    \includegraphics[width=0.99\linewidth]{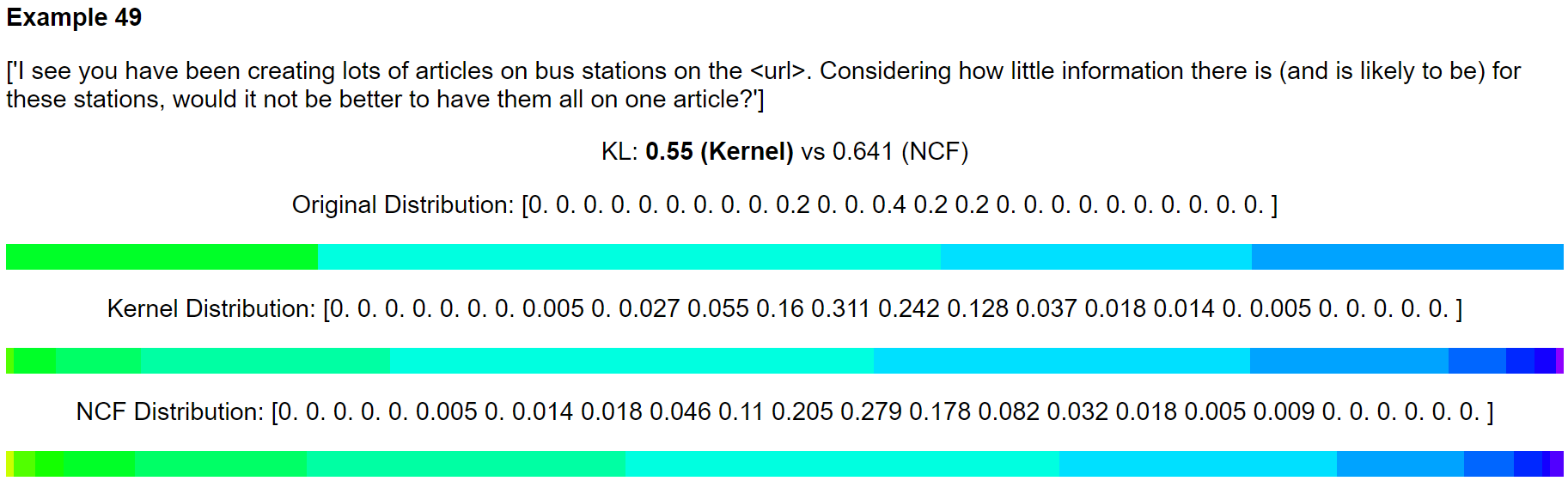}
    \includegraphics[width=0.99\linewidth]{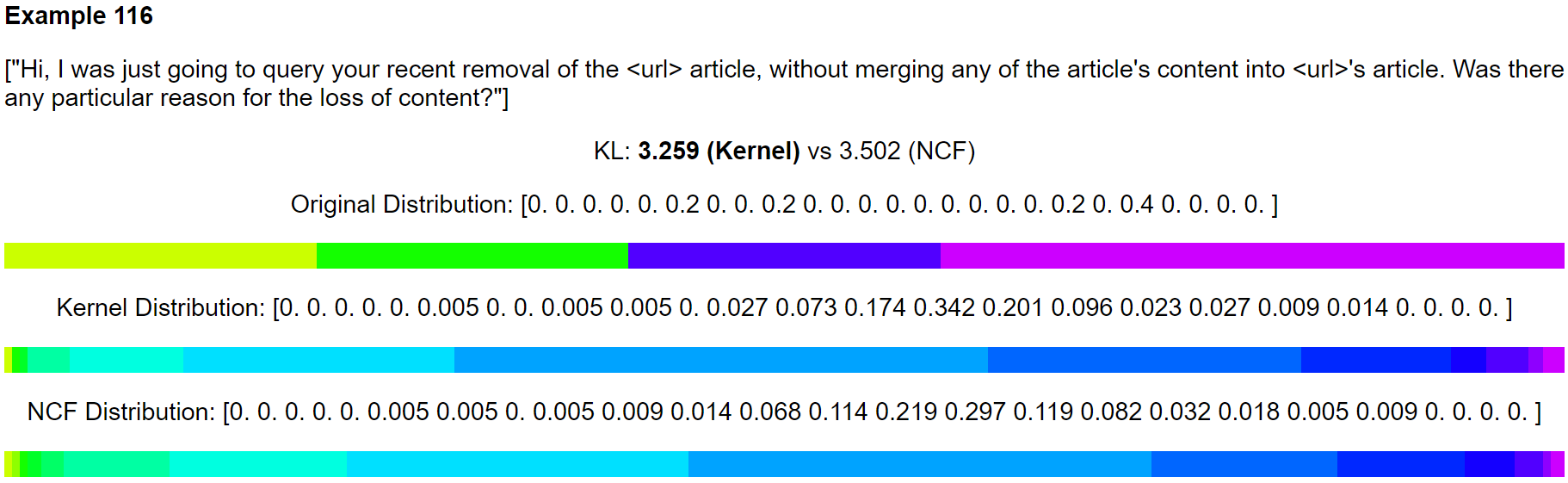}
    \includegraphics[width=0.99\linewidth]{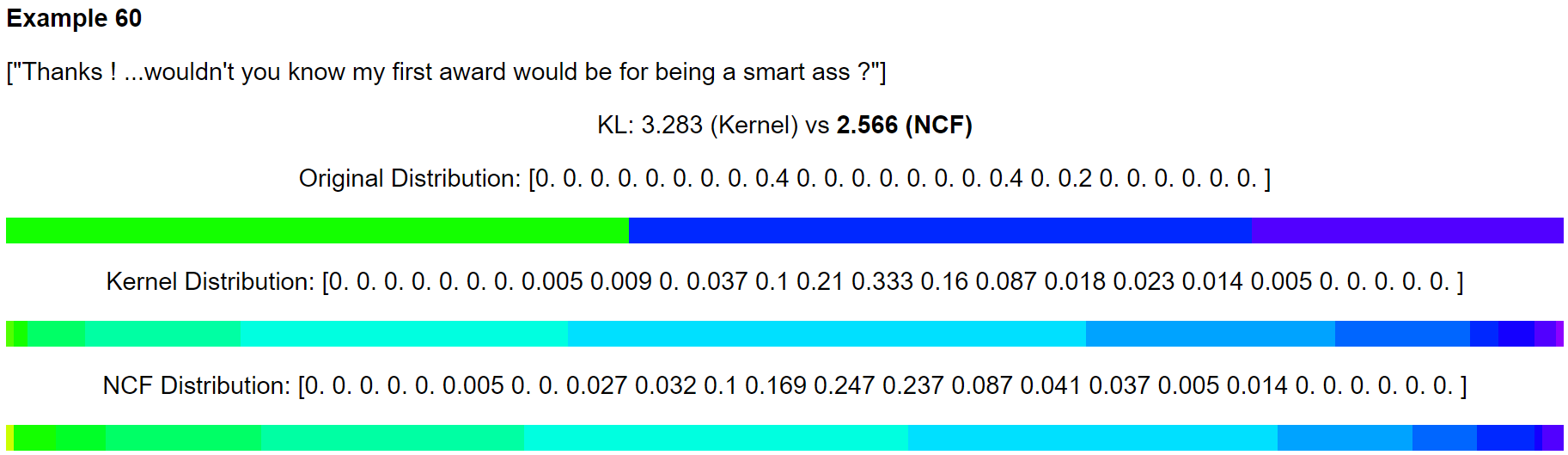}
    \caption{Examples from the Politeness dataset.}
    \label{fig:Politeness-website}
\end{figure*}

\begin{figure*}[t]
    \centering
    \includegraphics[width=0.99\linewidth]{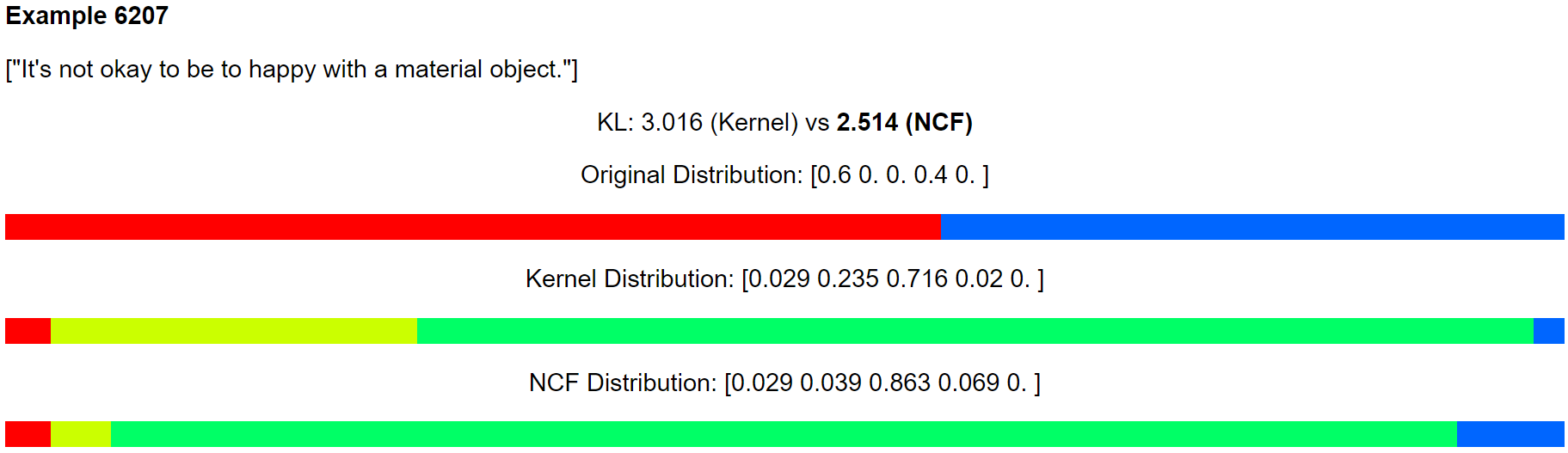}
    \includegraphics[width=0.99\linewidth]{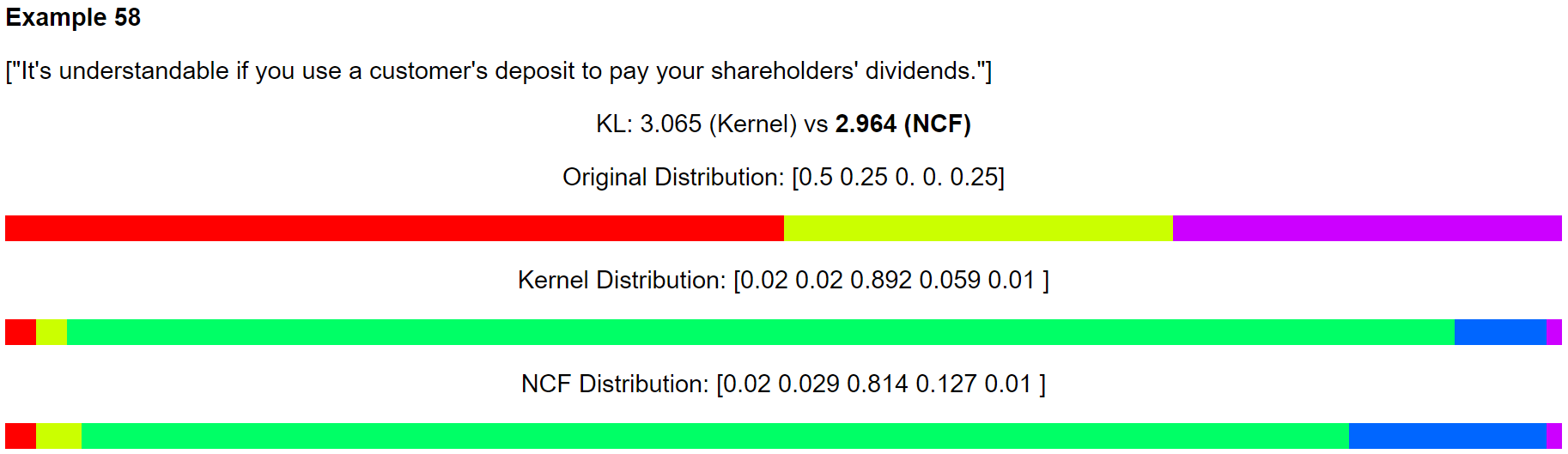}
    \includegraphics[width=0.99\linewidth]{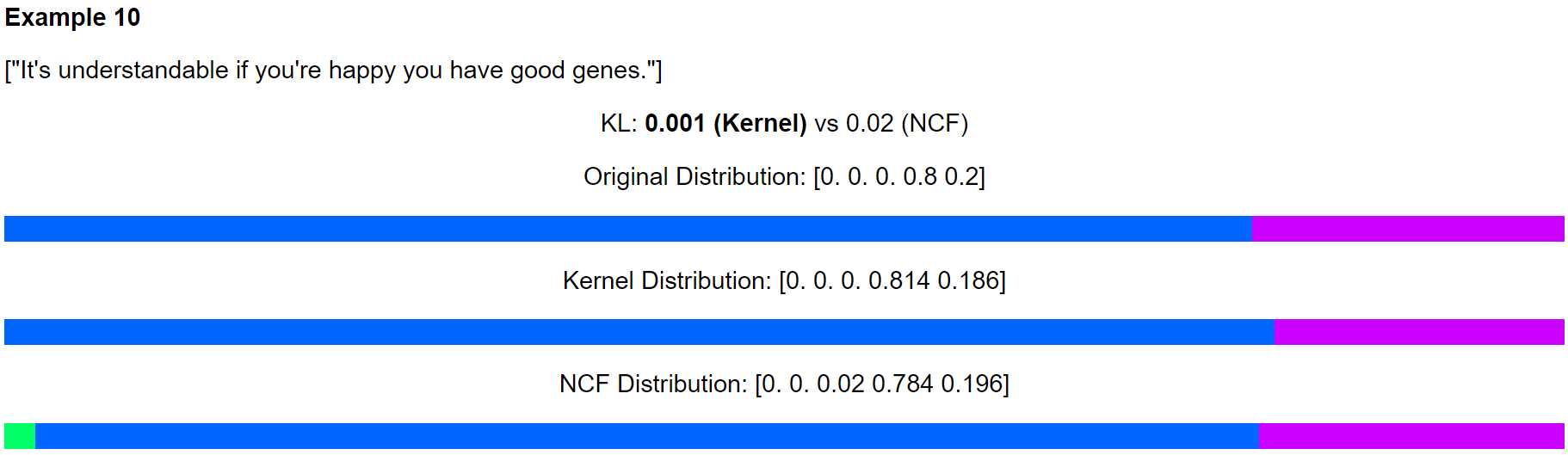}
    \includegraphics[width=0.99\linewidth]{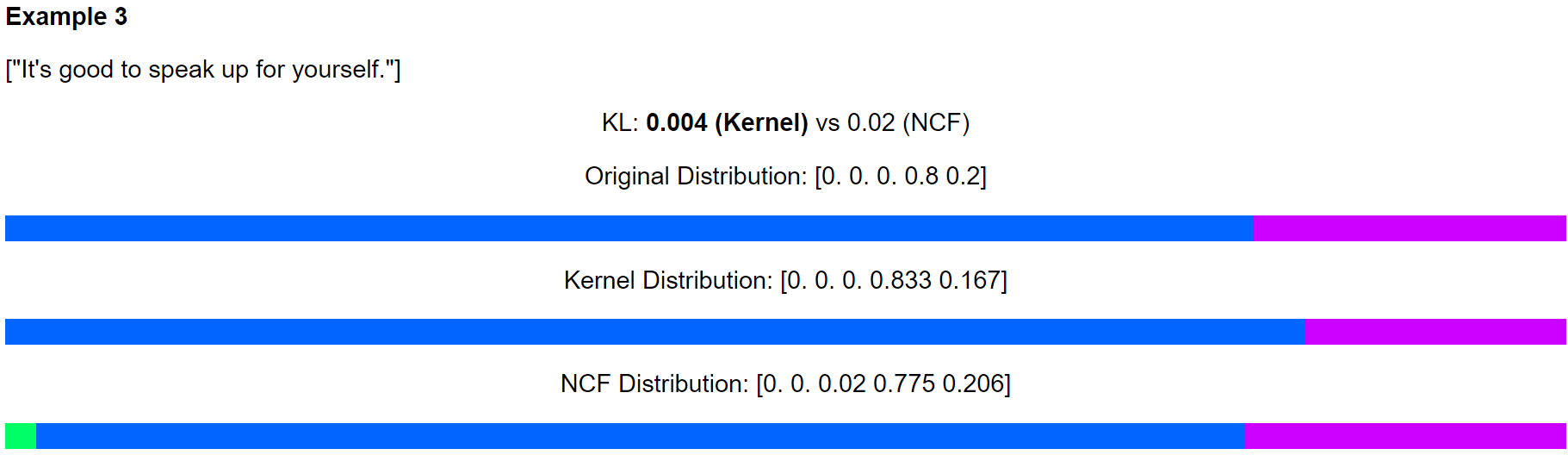}
    \caption{Examples from the SChem5Labels dataset.}
    \label{fig:SChem5Labels-website}
\end{figure*}

\begin{figure*}[t]
    \centering
    \includegraphics[width=0.99\linewidth]{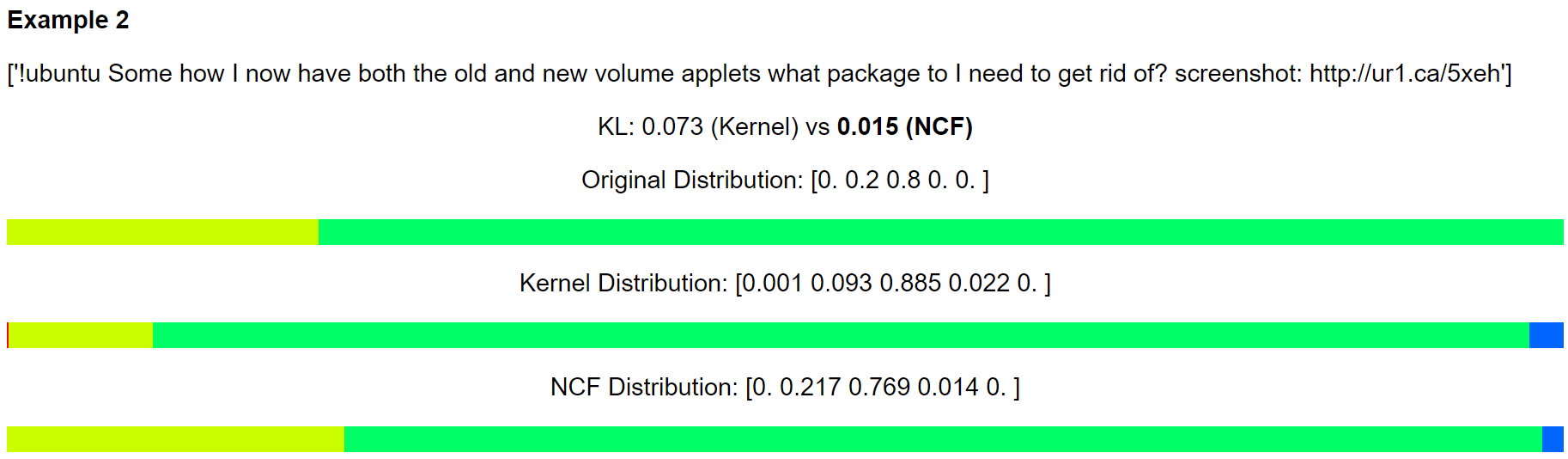}
    \includegraphics[width=0.99\linewidth]{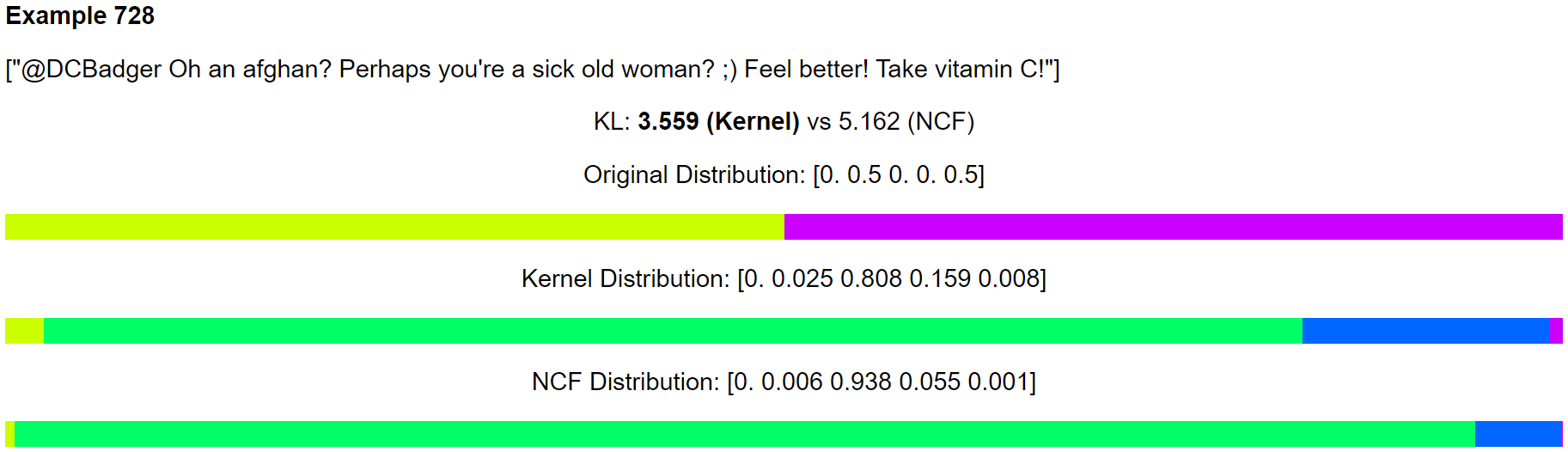}
    \includegraphics[width=0.99\linewidth]{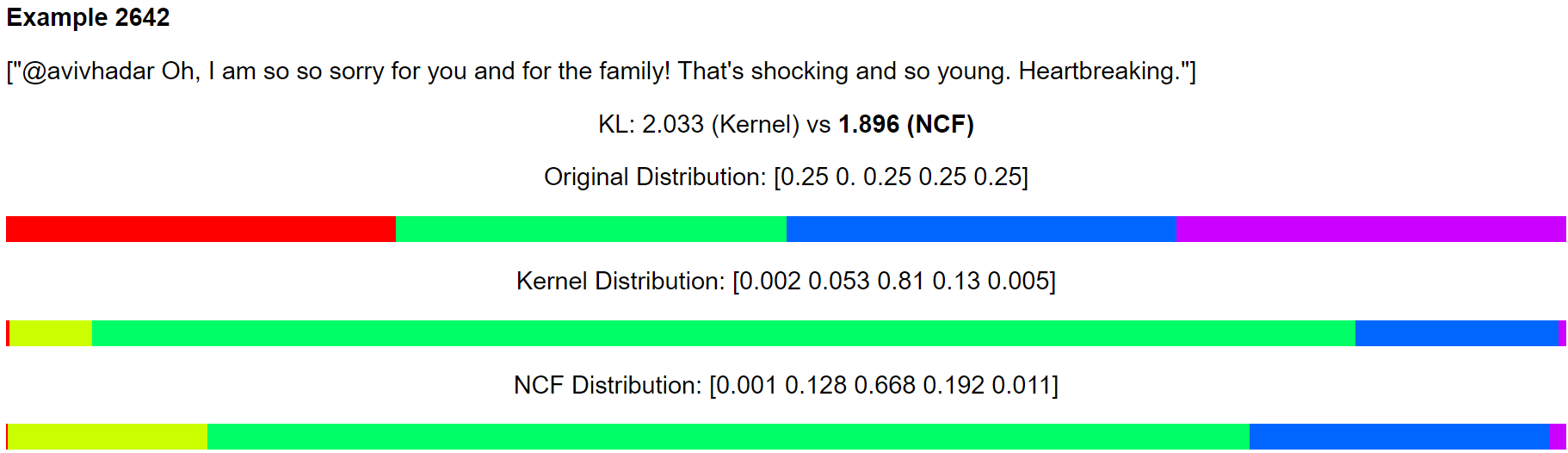}
    \includegraphics[width=0.99\linewidth]{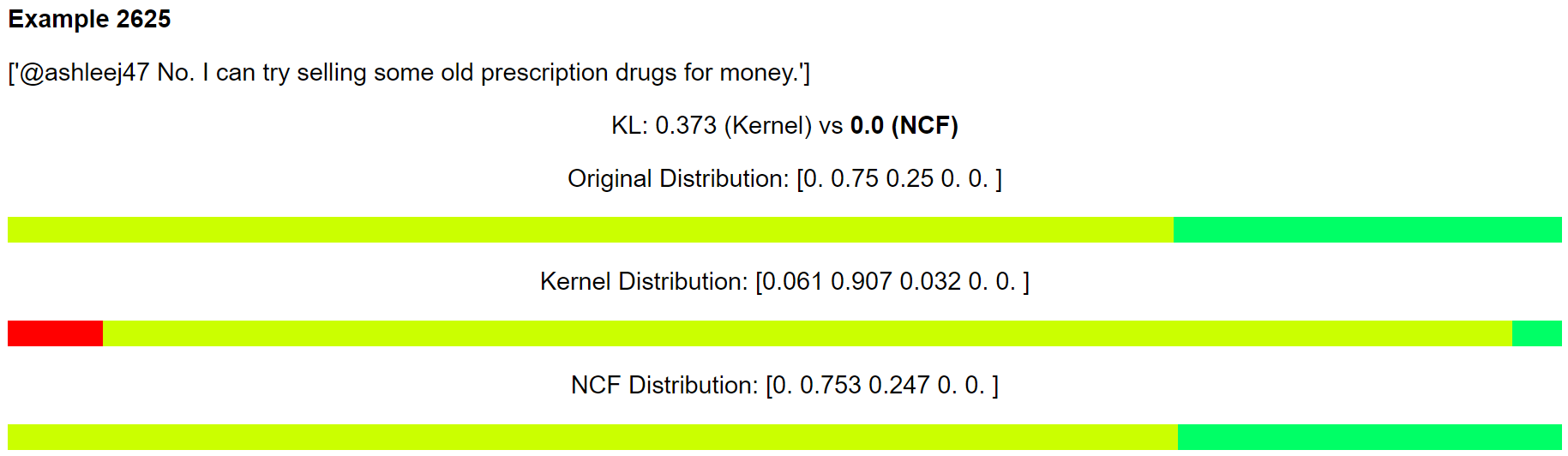}
    \caption{Examples from the Sentiment dataset.}
    \label{fig:Sentiment-website}
\end{figure*}

\begin{figure*}[h]
    \centering
    \includegraphics[width=0.99\linewidth]{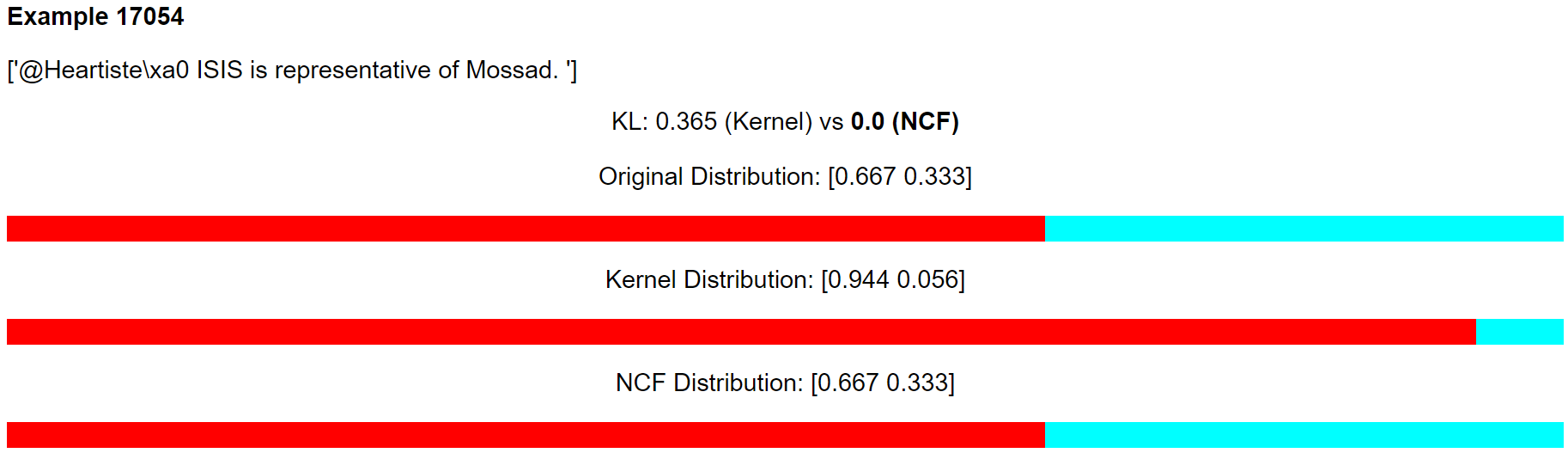}
    \includegraphics[width=0.99\linewidth]{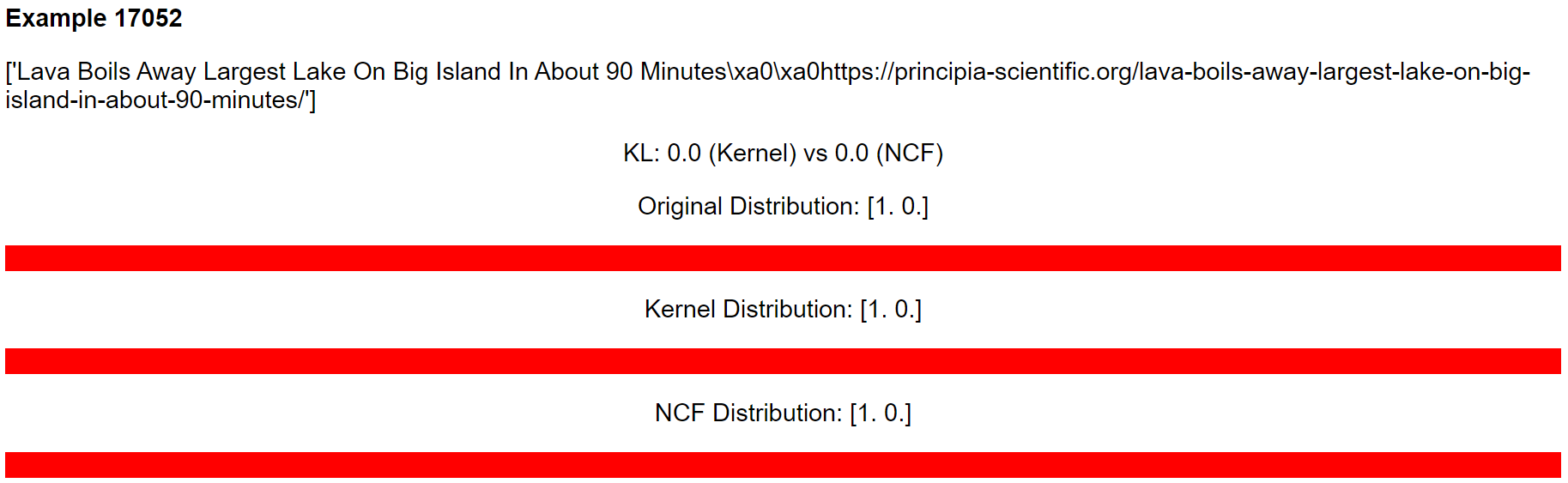}
    \includegraphics[width=0.99\linewidth]{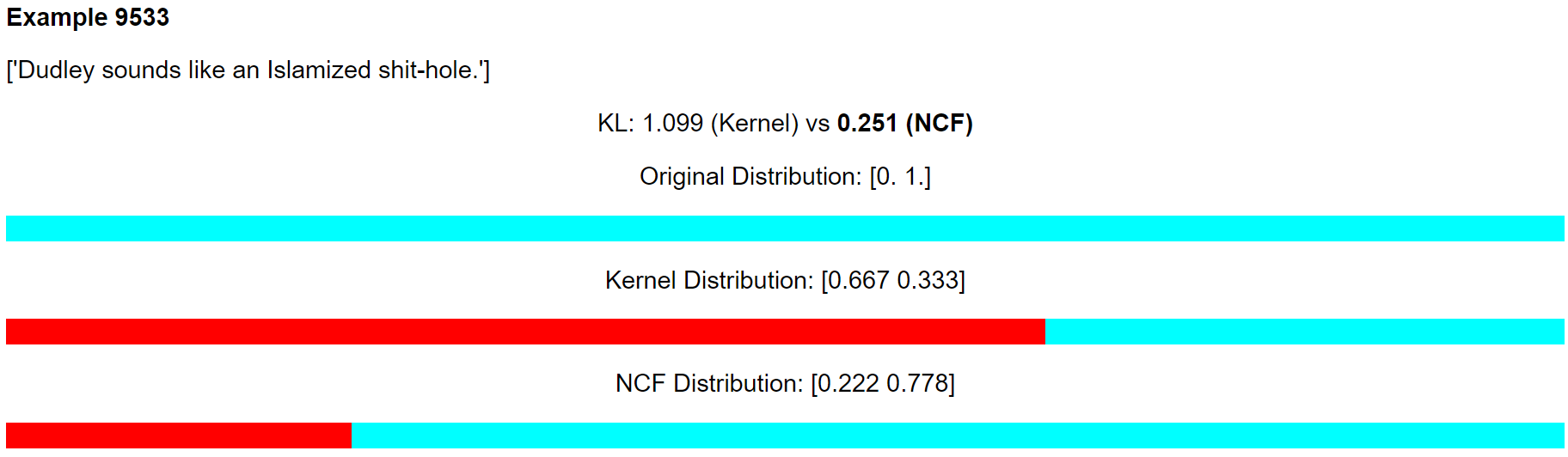}
    \caption{Examples from the GHC dataset.}
    \label{fig:GHC-website}
\end{figure*}

\begin{figure*}[h]
    \centering
    \includegraphics[width=0.99\linewidth]{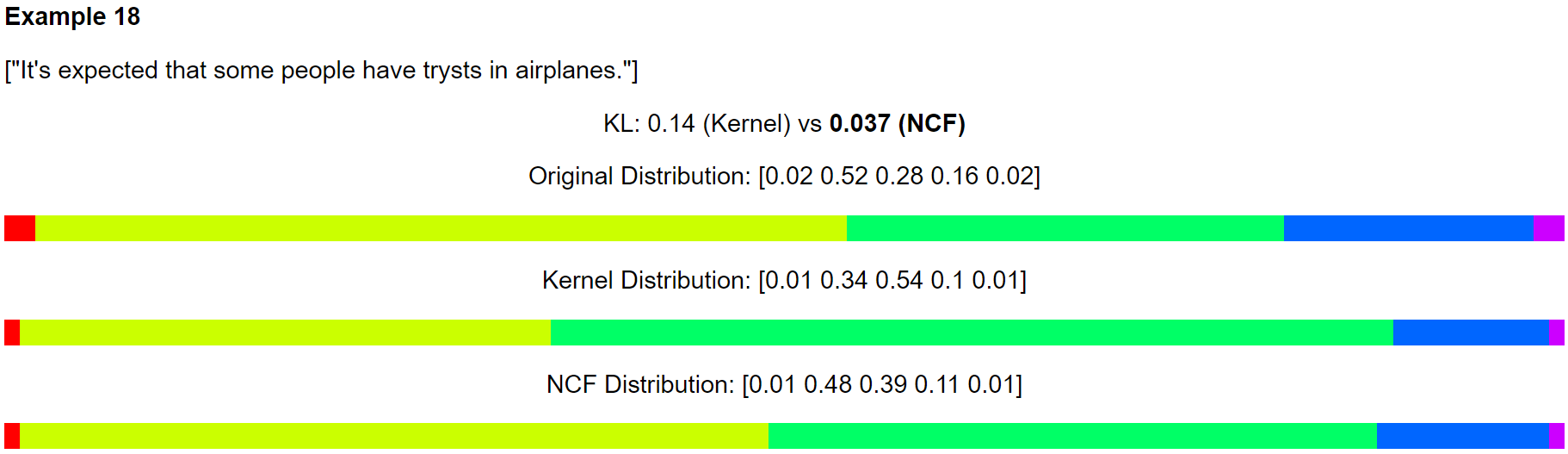}
    \includegraphics[width=0.99\linewidth]{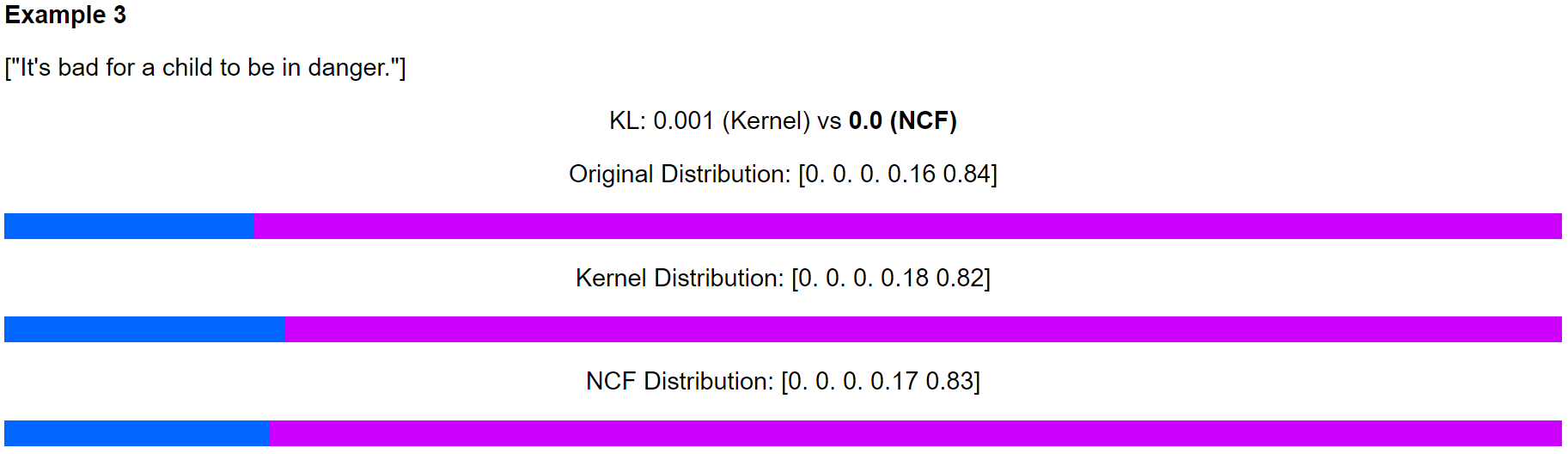}
    \includegraphics[width=0.99\linewidth]{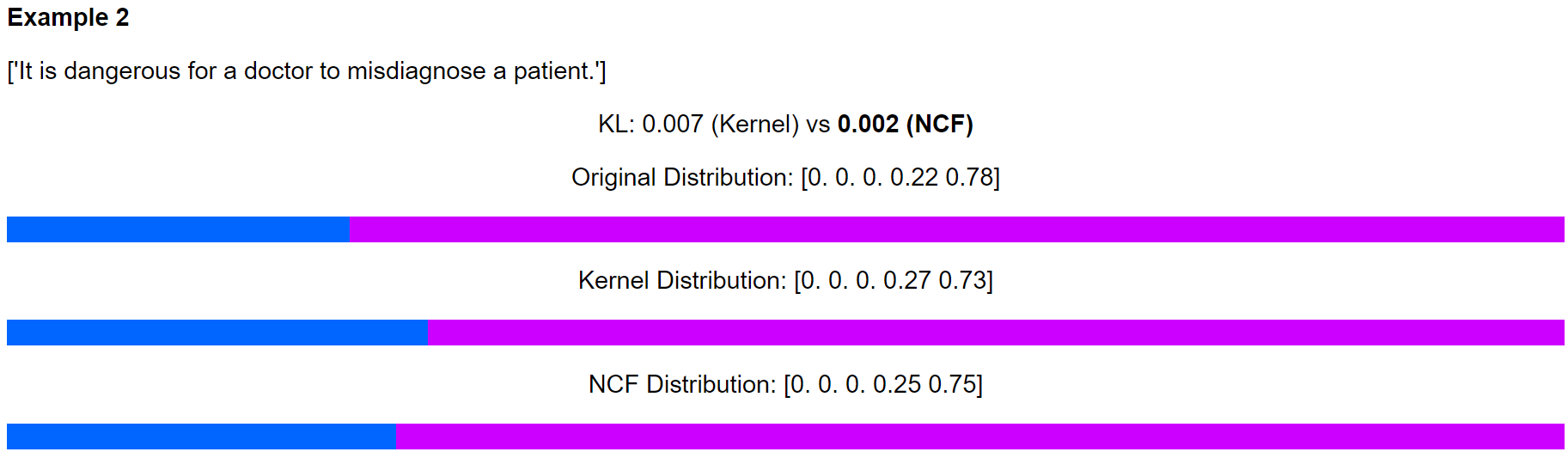}
    \caption{Additional examples from the SChem dataset.}
    \label{fig:SChem-website}
\end{figure*}

Our code to generate the full websites containing all of the examples is publicly available. Here, in Figures \ref{fig:SBIC-website}, \ref{fig:Politeness-website}, \ref{fig:SChem5Labels-website}, \ref{fig:Sentiment-website}, \ref{fig:GHC-website}, and \ref{fig:SChem-website}, we provide a subset of examples demonstrating high and low KL divergence scores from each of the datasets.

\section{Disagreement Levels}
\label{app:disagreement-levels}
The computation to determine whether an example has is ``low", ``medium", or ``high" disagreement was done individually for each fold of the data. When given a fold of data, we first compute the proportion of people who disagreed with the majority-voted label. (Note that ties in the majority-voted label do not impact this computation, since the same number of people will disagree regardless of which label is chosen among the tied options.) Then, we assign a threshold for ``low" and ``high" disagreement: any examples with disagreement equal to or lower than the ``low" threshold are considered to have ``low" disagreement, while any examples with disagreement equal to or greater than the ``high" threshold are considered to have ``high" disagreement. The number of examples in each category is a sum across all five folds of that dataset of examples that matched the threshold for that category.

The choice of thresholds must satisfy three rules: (1) The high threshold must be higher than the low threshold (2) There must be at least some examples in each category (3) The variance among the number of examples in each category must be minimized. When looking at Table \ref{tab:training_results_disagreement}, it may seem odd that the number of examples in each category is so varied, given the explicit minimization of variance in the rules. However, this occurs because there are many examples that have the exact same level of disagreement; rather than split these examples into two different categories, we opted to ensure that examples with the same level of disagreement were always labeled with the same level of disagreement.

\end{spverbatim}
\end{spverbatim}
\end{spverbatim}
\end{spverbatim}
\end{spverbatim}
\end{spverbatim}
\end{spverbatim}
\end{spverbatim}
\end{spverbatim}
\end{spverbatim}
\end{spverbatim}
\end{spverbatim}
\end{spverbatim}
\end{spverbatim}
\end{spverbatim}
\end{spverbatim}
\end{spverbatim}
\end{spverbatim}
\end{spverbatim}
\end{spverbatim}
\end{spverbatim}
\end{spverbatim}
\end{spverbatim}
\end{spverbatim}
\end{spverbatim}
\end{spverbatim}
\end{spverbatim}
\end{spverbatim}

\clearpage
\bibliography{ecai,anthology,custom,london}

\begin{thebibliography}{30}
\expandafter\ifx\csname natexlab\endcsname\relax\def\natexlab#1{#1}\fi
\providecommand{\url}[1]{\texttt{#1}}
\providecommand{\href}[2]{#2}
\providecommand{\path}[1]{#1}
\providecommand{\DOIprefix}{doi:}
\providecommand{\ArXivprefix}{arXiv:}
\providecommand{\URLprefix}{URL: }
\providecommand{\Pubmedprefix}{pmid:}
\providecommand{\doi}[1]{\href{http://dx.doi.org/#1}{\path{#1}}}
\providecommand{\Pubmed}[1]{\href{pmid:#1}{\path{#1}}}
\providecommand{\bibinfo}[2]{#2}
\ifx\xfnm\relax \def\xfnm[#1]{\unskip,\space#1}\fi
\bibitem[{Sharir et~al.(2020)Sharir, Peleg, and Shoham}]{sharir_cost_2020}
\bibinfo{author}{O.~Sharir}, \bibinfo{author}{B.~Peleg},
  \bibinfo{author}{Y.~Shoham}, \bibinfo{title}{The {Cost} of {Training} {NLP}
  {Models}: {A} {Concise} {Overview}}, \bibinfo{year}{2020}. \URLprefix
  \url{http://arxiv.org/abs/2004.08900}, \bibinfo{note}{arXiv:2004.08900 [cs]}.
\bibitem[{Checco et~al.(2017)Checco, Roitero, Maddalena, Mizzaro, and
  Demartini}]{checco2017let}
\bibinfo{author}{A.~Checco}, \bibinfo{author}{K.~Roitero},
  \bibinfo{author}{E.~Maddalena}, \bibinfo{author}{S.~Mizzaro},
  \bibinfo{author}{G.~Demartini},
\newblock \bibinfo{title}{Let's agree to disagree: Fixing agreement measures
  for crowdsourcing},
\newblock in: \bibinfo{booktitle}{Fifth AAAI Conference on Human Computation
  and Crowdsourcing}, \bibinfo{year}{2017}.
\bibitem[{Kairam and Heer(2016)}]{kairam2016parting}
\bibinfo{author}{S.~Kairam}, \bibinfo{author}{J.~Heer},
\newblock \bibinfo{title}{Parting crowds: Characterizing divergent
  interpretations in crowdsourced annotation tasks},
\newblock in: \bibinfo{booktitle}{Proceedings of the 19th ACM Conference on
  Computer-Supported Cooperative Work \& Social Computing},
  \bibinfo{year}{2016}, pp. \bibinfo{pages}{1637--1648}.
\bibitem[{Uma et~al.(2022)Uma, Almanea, and Poesio}]{Uma2022ScalingAD}
\bibinfo{author}{A.~Uma}, \bibinfo{author}{D.~Almanea},
  \bibinfo{author}{M.~Poesio},
\newblock \bibinfo{title}{Scaling and disagreements: Bias, noise, and
  ambiguity},
\newblock \bibinfo{journal}{Frontiers in Artificial Intelligence}
  \bibinfo{volume}{5} (\bibinfo{year}{2022}).
\bibitem[{Fornaciari et~al.(2021)Fornaciari, Uma, Paun, Plank, Hovy, and
  Poesio}]{fornaciari2021beyond}
\bibinfo{author}{T.~Fornaciari}, \bibinfo{author}{A.~Uma},
  \bibinfo{author}{S.~Paun}, \bibinfo{author}{B.~Plank},
  \bibinfo{author}{D.~Hovy}, \bibinfo{author}{M.~Poesio},
\newblock \bibinfo{title}{Beyond black \& white: Leveraging annotator
  disagreement via soft-label multi-task learning},
\newblock in: \bibinfo{booktitle}{2021 Conference of the North American Chapter
  of the Association for Computational Linguistics: Human Language
  Technologies}, \bibinfo{organization}{Association for Computational
  Linguistics}, \bibinfo{year}{2021}.
\bibitem[{Leonardelli et~al.(????)Leonardelli, Uma, Abercrombie, Almanea,
  Basile, Fornaciari, Plank, Rieser, and
  Poesio}]{leonardelli_semeval-2023_nodate}
\bibinfo{author}{E.~Leonardelli}, \bibinfo{author}{A.~Uma},
  \bibinfo{author}{G.~Abercrombie}, \bibinfo{author}{D.~Almanea},
  \bibinfo{author}{V.~Basile}, \bibinfo{author}{T.~Fornaciari},
  \bibinfo{author}{B.~Plank}, \bibinfo{author}{V.~Rieser},
  \bibinfo{author}{M.~Poesio},
\newblock \bibinfo{title}{{SemEval}-2023 {Task} 11: {Learning} {With}
  {Disagreements} ({LeWiDi})}  (????).
\bibitem[{Davani et~al.(2022)Davani, Díaz, and
  Prabhakaran}]{davani_dealing_2022}
\bibinfo{author}{A.~M. Davani}, \bibinfo{author}{M.~Díaz},
  \bibinfo{author}{V.~Prabhakaran},
\newblock \bibinfo{title}{Dealing with {Disagreements}: {Looking} {Beyond} the
  {Majority} {Vote} in {Subjective} {Annotations}},
\newblock \bibinfo{journal}{Transactions of the Association for Computational
  Linguistics} \bibinfo{volume}{10} (\bibinfo{year}{2022})
  \bibinfo{pages}{92--110}. \URLprefix
  \url{https://aclanthology.org/2022.tacl-1.6}.
  \DOIprefix\doi{10.1162/tacl_a_00449}, \bibinfo{note}{place: Cambridge, MA
  Publisher: MIT Press}.
\bibitem[{Rendle and Schmidt-Thieme(2008)}]{rendle_online-updating_2008}
\bibinfo{author}{S.~Rendle}, \bibinfo{author}{L.~Schmidt-Thieme},
\newblock \bibinfo{title}{Online-updating regularized kernel matrix
  factorization models for large-scale recommender systems},
\newblock in: \bibinfo{booktitle}{Proceedings of the 2008 {ACM} conference on
  {Recommender} systems}, {RecSys} '08, \bibinfo{publisher}{Association for
  Computing Machinery}, \bibinfo{address}{New York, NY, USA},
  \bibinfo{year}{2008}, pp. \bibinfo{pages}{251--258}. \URLprefix
  \url{https://doi.org/10.1145/1454008.1454047}.
  \DOIprefix\doi{10.1145/1454008.1454047}.
\bibitem[{He et~al.(2017)He, Liao, Zhang, Nie, Hu, and Chua}]{he_neural_2017}
\bibinfo{author}{X.~He}, \bibinfo{author}{L.~Liao}, \bibinfo{author}{H.~Zhang},
  \bibinfo{author}{L.~Nie}, \bibinfo{author}{X.~Hu}, \bibinfo{author}{T.-S.
  Chua},
\newblock \bibinfo{title}{Neural {Collaborative} {Filtering}},
\newblock in: \bibinfo{booktitle}{Proceedings of the 26th {International}
  {Conference} on {World} {Wide} {Web}}, {WWW} '17,
  \bibinfo{publisher}{International World Wide Web Conferences Steering
  Committee}, \bibinfo{address}{Republic and Canton of Geneva, CHE},
  \bibinfo{year}{2017}, pp. \bibinfo{pages}{173--182}. \URLprefix
  \url{https://doi.org/10.1145/3038912.3052569}.
  \DOIprefix\doi{10.1145/3038912.3052569}.
\bibitem[{Brown et~al.(2020)Brown, Mann, Ryder, Subbiah, Kaplan, Dhariwal,
  Neelakantan, Shyam, Sastry, Askell, Agarwal, Herbert-Voss, Krueger, Henighan,
  Child, Ramesh, Ziegler, Wu, Winter, Hesse, Chen, Sigler, Litwin, Gray, Chess,
  Clark, Berner, McCandlish, Radford, Sutskever, and
  Amodei}]{brown_language_2020}
\bibinfo{author}{T.~Brown}, \bibinfo{author}{B.~Mann},
  \bibinfo{author}{N.~Ryder}, \bibinfo{author}{M.~Subbiah},
  \bibinfo{author}{J.~D. Kaplan}, \bibinfo{author}{P.~Dhariwal},
  \bibinfo{author}{A.~Neelakantan}, \bibinfo{author}{P.~Shyam},
  \bibinfo{author}{G.~Sastry}, \bibinfo{author}{A.~Askell},
  \bibinfo{author}{S.~Agarwal}, \bibinfo{author}{A.~Herbert-Voss},
  \bibinfo{author}{G.~Krueger}, \bibinfo{author}{T.~Henighan},
  \bibinfo{author}{R.~Child}, \bibinfo{author}{A.~Ramesh},
  \bibinfo{author}{D.~Ziegler}, \bibinfo{author}{J.~Wu},
  \bibinfo{author}{C.~Winter}, \bibinfo{author}{C.~Hesse},
  \bibinfo{author}{M.~Chen}, \bibinfo{author}{E.~Sigler},
  \bibinfo{author}{M.~Litwin}, \bibinfo{author}{S.~Gray},
  \bibinfo{author}{B.~Chess}, \bibinfo{author}{J.~Clark},
  \bibinfo{author}{C.~Berner}, \bibinfo{author}{S.~McCandlish},
  \bibinfo{author}{A.~Radford}, \bibinfo{author}{I.~Sutskever},
  \bibinfo{author}{D.~Amodei},
\newblock \bibinfo{title}{Language {Models} are {Few}-{Shot} {Learners}},
\newblock in: \bibinfo{booktitle}{Advances in {Neural} {Information}
  {Processing} {Systems}}, volume~\bibinfo{volume}{33},
  \bibinfo{publisher}{Curran Associates, Inc.}, \bibinfo{year}{2020}, pp.
  \bibinfo{pages}{1877--1901}. \URLprefix
  \url{https://proceedings.neurips.cc/paper/2020/hash/1457c0d6bfcb4967418bfb8ac142f64a-Abstract.html}.
\bibitem[{Poesio and Artstein(2005)}]{poesio2005reliability}
\bibinfo{author}{M.~Poesio}, \bibinfo{author}{R.~Artstein},
\newblock \bibinfo{title}{The reliability of anaphoric annotation,
  reconsidered: Taking ambiguity into account},
\newblock in: \bibinfo{booktitle}{Proceedings of the workshop on frontiers in
  corpus annotations ii: Pie in the sky}, \bibinfo{year}{2005}, pp.
  \bibinfo{pages}{76--83}.
\bibitem[{Versley(2008)}]{versley_vagueness_2008}
\bibinfo{author}{Y.~Versley},
\newblock \bibinfo{title}{Vagueness and {Referential} {Ambiguity} in a
  {Large}-{Scale} {Annotated} {Corpus}},
\newblock \bibinfo{journal}{Research on Language and Computation}
  \bibinfo{volume}{6} (\bibinfo{year}{2008}) \bibinfo{pages}{333--353}.
  \URLprefix \url{https://doi.org/10.1007/s11168-008-9059-1}.
  \DOIprefix\doi{10.1007/s11168-008-9059-1}.
\bibitem[{Wan and Badillo-Urquiola(2023)}]{wan-badillo-urquiola-2023-dragonfly}
\bibinfo{author}{R.~Wan}, \bibinfo{author}{K.~Badillo-Urquiola},
\newblock \bibinfo{title}{Dragonfly{\_}captain at {S}em{E}val-2023 task 11:
  Unpacking disagreement with investigation of annotator demographics and task
  difficulty},
\newblock in: \bibinfo{booktitle}{Proceedings of the 17th International
  Workshop on Semantic Evaluation (SemEval-2023)},
  \bibinfo{publisher}{Association for Computational Linguistics},
  \bibinfo{address}{Toronto, Canada}, \bibinfo{year}{2023}, pp.
  \bibinfo{pages}{1978--1982}. \URLprefix
  \url{https://aclanthology.org/2023.semeval-1.272}.
  \DOIprefix\doi{10.18653/v1/2023.semeval-1.272}.
\bibitem[{Gordon et~al.(2022)Gordon, Lam, Park, Patel, Hancock, Hashimoto, and
  Bernstein}]{gordon_jury_2022}
\bibinfo{author}{M.~L. Gordon}, \bibinfo{author}{M.~S. Lam},
  \bibinfo{author}{J.~S. Park}, \bibinfo{author}{K.~Patel},
  \bibinfo{author}{J.~T. Hancock}, \bibinfo{author}{T.~Hashimoto},
  \bibinfo{author}{M.~S. Bernstein},
\newblock \bibinfo{title}{Jury {Learning}: {Integrating} {Dissenting} {Voices}
  into {Machine} {Learning} {Models}},
\newblock \bibinfo{year}{2022}. \URLprefix
  \url{http://arxiv.org/abs/2202.02950}.
  \DOIprefix\doi{10.1145/3491102.3502004}, \bibinfo{note}{arXiv:2202.02950
  [cs]}.
\bibitem[{Gordon et~al.(2021)Gordon, Zhou, Patel, Hashimoto, and
  Bernstein}]{gordon_disagreement_2021}
\bibinfo{author}{M.~L. Gordon}, \bibinfo{author}{K.~Zhou},
  \bibinfo{author}{K.~Patel}, \bibinfo{author}{T.~Hashimoto},
  \bibinfo{author}{M.~S. Bernstein},
\newblock \bibinfo{title}{The {Disagreement} {Deconvolution}: {Bringing}
  {Machine} {Learning} {Performance} {Metrics} {In} {Line} {With} {Reality}},
\newblock in: \bibinfo{booktitle}{Proceedings of the 2021 {CHI} {Conference} on
  {Human} {Factors} in {Computing} {Systems}}, \bibinfo{publisher}{ACM},
  \bibinfo{address}{Yokohama Japan}, \bibinfo{year}{2021}, pp.
  \bibinfo{pages}{1--14}. \URLprefix
  \url{https://dl.acm.org/doi/10.1145/3411764.3445423}.
  \DOIprefix\doi{10.1145/3411764.3445423}.
\bibitem[{Wan et~al.(2023)Wan, Kim, and Kang}]{wan2023everyone}
\bibinfo{author}{R.~Wan}, \bibinfo{author}{J.~Kim}, \bibinfo{author}{D.~Kang},
\newblock \bibinfo{title}{Everyone's voice matters: Quantifying annotation
  disagreement using demographic information},
\newblock \bibinfo{journal}{arXiv preprint arXiv:2301.05036}
  (\bibinfo{year}{2023}).
\bibitem[{Uma et~al.(2021)Uma, Fornaciari, Hovy, Paun, Plank, and
  Poesio}]{uma_learning_2021}
\bibinfo{author}{A.~N. Uma}, \bibinfo{author}{T.~Fornaciari},
  \bibinfo{author}{D.~Hovy}, \bibinfo{author}{S.~Paun},
  \bibinfo{author}{B.~Plank}, \bibinfo{author}{M.~Poesio},
\newblock \bibinfo{title}{Learning from {Disagreement}: {A} {Survey}},
\newblock \bibinfo{journal}{Journal of Artificial Intelligence Research}
  \bibinfo{volume}{72} (\bibinfo{year}{2021}) \bibinfo{pages}{1385--1470}.
  \URLprefix \url{https://www.jair.org/index.php/jair/article/view/12752}.
  \DOIprefix\doi{10.1613/jair.1.12752}.
\bibitem[{Herlocker et~al.(2004)Herlocker, Konstan, Terveen, and
  Riedl}]{herlocker_evaluating_2004}
\bibinfo{author}{J.~L. Herlocker}, \bibinfo{author}{J.~A. Konstan},
  \bibinfo{author}{L.~G. Terveen}, \bibinfo{author}{J.~T. Riedl},
\newblock \bibinfo{title}{Evaluating collaborative filtering recommender
  systems},
\newblock \bibinfo{journal}{ACM Transactions on Information Systems}
  \bibinfo{volume}{22} (\bibinfo{year}{2004}) \bibinfo{pages}{5--53}.
  \URLprefix \url{https://dl.acm.org/doi/10.1145/963770.963772}.
  \DOIprefix\doi{10.1145/963770.963772}.
\bibitem[{Isinkaye et~al.(2015)Isinkaye, Folajimi, and
  Ojokoh}]{isinkaye_recommendation_2015}
\bibinfo{author}{F.~O. Isinkaye}, \bibinfo{author}{Y.~O. Folajimi},
  \bibinfo{author}{B.~A. Ojokoh},
\newblock \bibinfo{title}{Recommendation systems: {Principles}, methods and
  evaluation},
\newblock \bibinfo{journal}{Egyptian Informatics Journal} \bibinfo{volume}{16}
  (\bibinfo{year}{2015}) \bibinfo{pages}{261--273}. \URLprefix
  \url{https://www.sciencedirect.com/science/article/pii/S1110866515000341}.
  \DOIprefix\doi{10.1016/j.eij.2015.06.005}.
\bibitem[{Do(2022)}]{do_matrix_2022}
\bibinfo{author}{Q.-V. Do}, \bibinfo{title}{Matrix {Factorization}},
  \bibinfo{year}{2022}. \URLprefix
  \url{https://github.com/Quang-Vinh/matrix-factorization},
  \bibinfo{note}{original-date: 2020-06-04T00:10:11Z}.
\bibitem[{Forbes et~al.(2020)Forbes, Hwang, Shwartz, Sap, and
  Choi}]{Forbes2020SocialC1}
\bibinfo{author}{M.~Forbes}, \bibinfo{author}{J.~D. Hwang},
  \bibinfo{author}{V.~Shwartz}, \bibinfo{author}{M.~Sap},
  \bibinfo{author}{Y.~Choi},
\newblock \bibinfo{title}{Social chemistry 101: Learning to reason about social
  and moral norms},
\newblock \bibinfo{journal}{ArXiv} \bibinfo{volume}{abs/2011.00620}
  (\bibinfo{year}{2020}).
\bibitem[{Sap et~al.(2020)Sap, Gabriel, Qin, Jurafsky, Smith, and
  Choi}]{Sap2020SocialBF}
\bibinfo{author}{M.~Sap}, \bibinfo{author}{S.~Gabriel},
  \bibinfo{author}{L.~Qin}, \bibinfo{author}{D.~Jurafsky},
  \bibinfo{author}{N.~A. Smith}, \bibinfo{author}{Y.~Choi},
\newblock \bibinfo{title}{Social bias frames: Reasoning about social and power
  implications of language},
\newblock \bibinfo{journal}{ArXiv} \bibinfo{volume}{abs/1911.03891}
  (\bibinfo{year}{2020}).
\bibitem[{Kennedy et~al.(2018)Kennedy, Atari, Davani, Yeh, Omrani, Kim, Coombs,
  Havaldar, Portillo-Wightman, Gonzalez et~al.}]{kennedy2018introducing}
\bibinfo{author}{B.~Kennedy}, \bibinfo{author}{M.~Atari},
  \bibinfo{author}{A.~Davani}, \bibinfo{author}{L.~Yeh},
  \bibinfo{author}{A.~Omrani}, \bibinfo{author}{Y.~Kim},
  \bibinfo{author}{K.~Coombs}, \bibinfo{author}{S.~Havaldar},
  \bibinfo{author}{G.~Portillo-Wightman}, \bibinfo{author}{E.~Gonzalez},
  et~al.,
\newblock \bibinfo{title}{Introducing the gab hate corpus: Defining and
  applying hate-based rhetoric to social media posts at scale}
  (\bibinfo{year}{2018}).
\bibitem[{Kennedy et~al.(2022)Kennedy, Atari, Davani, Yeh, Omrani, Kim, Coombs,
  Havaldar, Portillo-Wightman, Gonzalez et~al.}]{kennedy2022introducing}
\bibinfo{author}{B.~Kennedy}, \bibinfo{author}{M.~Atari},
  \bibinfo{author}{A.~M. Davani}, \bibinfo{author}{L.~Yeh},
  \bibinfo{author}{A.~Omrani}, \bibinfo{author}{Y.~Kim},
  \bibinfo{author}{K.~Coombs}, \bibinfo{author}{S.~Havaldar},
  \bibinfo{author}{G.~Portillo-Wightman}, \bibinfo{author}{E.~Gonzalez},
  et~al.,
\newblock \bibinfo{title}{Introducing the gab hate corpus: defining and
  applying hate-based rhetoric to social media posts at scale},
\newblock \bibinfo{journal}{Language Resources and Evaluation}
  \bibinfo{volume}{56} (\bibinfo{year}{2022}) \bibinfo{pages}{79--108}.
\bibitem[{Diaz et~al.(2018)Diaz, Johnson, Lazar, Piper, and
  Gergle}]{Diaz2018AddressingAB}
\bibinfo{author}{M.~Diaz}, \bibinfo{author}{I.~L. Johnson},
  \bibinfo{author}{A.~Lazar}, \bibinfo{author}{A.~M. Piper},
  \bibinfo{author}{D.~Gergle},
\newblock \bibinfo{title}{Addressing age-related bias in sentiment analysis},
\newblock \bibinfo{journal}{Proceedings of the 2018 CHI Conference on Human
  Factors in Computing Systems}  (\bibinfo{year}{2018}).
\bibitem[{Danescu-Niculescu-Mizil et~al.(2013)Danescu-Niculescu-Mizil, Sudhof,
  Jurafsky, Leskovec, and Potts}]{DanescuNiculescuMizil2013ACA}
\bibinfo{author}{C.~Danescu-Niculescu-Mizil}, \bibinfo{author}{M.~Sudhof},
  \bibinfo{author}{D.~Jurafsky}, \bibinfo{author}{J.~Leskovec},
  \bibinfo{author}{C.~Potts},
\newblock \bibinfo{title}{A computational approach to politeness with
  application to social factors},
\newblock in: \bibinfo{booktitle}{ACL}, \bibinfo{year}{2013}.
\bibitem[{Roy(2019)}]{roy_all_2019}
\bibinfo{author}{B.~Roy}, \bibinfo{title}{All {About} {Missing} {Data}
  {Handling}. {Missing} data is a every day problem… {\textbar} by
  {Baijayanta} {Roy} {\textbar} {Towards} {Data} {Science}},
  \bibinfo{year}{2019}. \URLprefix
  \url{https://towardsdatascience.com/all-about-missing-data-handling-b94b8b5d2184}.
\bibitem[{Kaplan et~al.(2020)Kaplan, McCandlish, Henighan, Brown, Chess, Child,
  Gray, Radford, Wu, and Amodei}]{kaplan_scaling_2020}
\bibinfo{author}{J.~Kaplan}, \bibinfo{author}{S.~McCandlish},
  \bibinfo{author}{T.~Henighan}, \bibinfo{author}{T.~B. Brown},
  \bibinfo{author}{B.~Chess}, \bibinfo{author}{R.~Child},
  \bibinfo{author}{S.~Gray}, \bibinfo{author}{A.~Radford},
  \bibinfo{author}{J.~Wu}, \bibinfo{author}{D.~Amodei}, \bibinfo{title}{Scaling
  {Laws} for {Neural} {Language} {Models}}, \bibinfo{year}{2020}. \URLprefix
  \url{http://arxiv.org/abs/2001.08361}.
  \DOIprefix\doi{10.48550/arXiv.2001.08361}, \bibinfo{note}{arXiv:2001.08361
  [cs, stat]}.
\bibitem[{Devlin et~al.(2019)Devlin, Chang, Lee, and
  Toutanova}]{devlin_bert_2019}
\bibinfo{author}{J.~Devlin}, \bibinfo{author}{M.-W. Chang},
  \bibinfo{author}{K.~Lee}, \bibinfo{author}{K.~Toutanova},
\newblock \bibinfo{title}{{BERT}: {Pre}-training of {Deep} {Bidirectional}
  {Transformers} for {Language} {Understanding}},
\newblock in: \bibinfo{booktitle}{Proceedings of the 2019 {Conference} of the
  {North} {American} {Chapter} of the {Association} for {Computational}
  {Linguistics}: {Human} {Language} {Technologies}, {Volume} 1 ({Long} and
  {Short} {Papers})}, \bibinfo{publisher}{Association for Computational
  Linguistics}, \bibinfo{address}{Minneapolis, Minnesota},
  \bibinfo{year}{2019}, pp. \bibinfo{pages}{4171--4186}. \URLprefix
  \url{https://aclanthology.org/N19-1423}.
  \DOIprefix\doi{10.18653/v1/N19-1423}.
\bibitem[{~(2023)}]{noauthor_bert-base-uncased_2023}
\bibinfo{author}{H.~F. ~}, \bibinfo{title}{bert-base-uncased · {Hugging}
  {Face}}, \bibinfo{year}{2023}. \URLprefix
  \url{https://huggingface.co/bert-base-uncased}.

\end{thebibliography}

\end{document}